\pdfoutput=1

\documentclass[10pt,journal,compsoc]{IEEEtran}

\ifCLASSOPTIONcompsoc
  \usepackage[nocompress]{cite}
\else
  \usepackage{cite}
\fi

\ifCLASSINFOpdf
\else
\fi

\usepackage[x11names,usenames,dvipsnames]{xcolor}
\usepackage{forest}
\usepackage{tikz}
\usepackage{graphics}
\usepackage{pgfplots}
\pgfplotsset{compat=1.17}
\usepackage{newtxmath}

\usepackage{graphicx}

\usepackage{algorithm}

\usepackage{algorithmic}
\usepackage{tabularx}
\usepackage[backref]{hyperref}
\hypersetup{
    colorlinks=true, 
    linkcolor=black, 
    citecolor=black, 
    filecolor=black, 
    urlcolor=black, 
    pdfnewwindow=false, 
    linktoc=none 
}
\usepackage{etoolbox}
\AtBeginEnvironment{IEEEbiography}{\vspace{-2.6\baselineskip}}
\usepackage{float}
\usepackage{multirow}
\usepackage{url}
\usepackage[numbers,sort&compress]{natbib}
\usepackage{amsmath,amsfonts,bm,amssymb,mathtools,amsthm}
\usepackage{color,xcolor}
\usepackage{amsmath}
\usepackage{xspace}

\def\Secref#1{Section~\ref{#1}}

\def\eqref#1{equation~\ref{#1}}
\def\Eqref#1{Eq.~(\ref{#1})}

\def\1{\bm{1}}

\DeclareMathAlphabet{\mathsfit}{\encodingdefault}{\sfdefault}{m}{sl}
\SetMathAlphabet{\mathsfit}{bold}{\encodingdefault}{\sfdefault}{bx}{n}
\begin{document}

\title{A Survey on Generative Diffusion Models}

\author{Hanqun~Cao,
        Cheng~Tan,
        Zhangyang~Gao,
        Yilun~Xu,
        Guangyong~Chen,
        Pheng-Ann~Heng,~\IEEEmembership{Senior Member,~IEEE},
        and~Stan~Z.~Li,~\IEEEmembership{Fellow,~IEEE}
\IEEEcompsocitemizethanks{
\IEEEcompsocthanksitem H. Cao is with the Department of Computer Science and Engineering, The Chinese University of Hong Kong, Hong Kong, China, also with Zhejiang Lab, Hangzhou, China. Email: 1155141481@link.cuhk.edu.hk.
\IEEEcompsocthanksitem C. Tan and Z. Gao are with the AI Lab, School of Engineering, Westlake University, Hangzhou, China. Email: {tancheng, gaozhangyang}@westlake.edu.cn.
\IEEEcompsocthanksitem Y. Xu is with Massachusetts Institute of Technology, Cambridge, Massachusetts, U.S. Email: ylxu@mit.edu.
\IEEEcompsocthanksitem G. Chen is with Zhejiang Lab, Hangzhou, China. Email: gychen@zhejianglab.com. 
\IEEEcompsocthanksitem P.-A. Heng is with the Department of Computer Science and Engineering, The Chinese University of Hong Kong, Hong Kong, China.
\IEEEcompsocthanksitem Stan Z. Li is with the AI Lab, School of Engineering, Westlake University, Hangzhou, China. Email: Stan.ZQ.Li@westlake.edu.cn. 
\IEEEcompsocthanksitem H. Cao, C. Tan, and Z. Gao contributed equally to this work. \protect\\
}
}

\markboth{Journal of \LaTeX\ Class Files,}%
{Shell \MakeLowercase{\textit{\textit{et al.}}}: Bare Demo of IEEEtran.cls for Computer Society Journals}

\IEEEtitleabstractindextext{
\begin{abstract}
Deep generative models have unlocked another profound realm of human creativity. By capturing and generalizing patterns within data, we have entered the epoch of all-encompassing Artificial Intelligence for General Creativity (AIGC). Notably, diffusion models, recognized as one of the paramount generative models, materialize human ideation into tangible instances across diverse domains, encompassing imagery, text, speech, biology, and healthcare. To provide advanced and comprehensive insights into diffusion, this survey comprehensively elucidates its developmental trajectory and future directions from three distinct angles: the fundamental formulation of diffusion, algorithmic enhancements, and the manifold applications of diffusion. Each layer is meticulously explored to offer a profound comprehension of its evolution. Structured and summarized approaches are presented \href{https://github.com/chq1155/A-Survey-on-Generative-Diffusion-Model}{here}.
\end{abstract}

\begin{IEEEkeywords}
Diffusion Model, Deep Generative Model, Diffusion Algorithm, Diffusion Applications.
\end{IEEEkeywords}}

\maketitle
\IEEEdisplaynontitleabstractindextext
\IEEEpeerreviewmaketitle

\IEEEraisesectionheading{\section{Introduction}}
\label{sec:introduction}
\IEEEPARstart{H}ow can we enable machines to possess human-like imagination? Deep generative models, including Variational Autoencoders (VAEs) \cite{kingma2019introduction, oussidi2018deep}, Energy-Based Models (EBMs) \cite{lecun2006tutorial, ngiam2011learning}, Generative Adversarial Networks (GANs) \cite{goodfellow2020generative, creswell2018generative}, normalizing flows (NFs) \cite{rezende2015variational, kobyzev2020normalizing}, and diffusion models \cite{sohl2015deep, ho2020denoising, song2020score}, have demonstrated remarkable potential in generating realistic samples. Within this survey, our central emphasis lies on diffusion models, which epitomize the forefront of advancements within this domain. These models effectively surmount the obstacles entailed in aligning posterior distributions within VAEs, mitigating the instability inherent in adversarial objectives of GANs, addressing the computational burdens associated with Markov Chain Monte Carlo (MCMC) methods during training in EBMs, and enforcing network constraints akin to NFs. Consequently, diffusion models have garnered significant attention in various domains, including computer vision~\cite{ho2022video, batzolis2021conditional,Song2023ConsistencyM}, natural language processing~\cite{li2022diffusion,austin2021structured}, time series~\cite{tashiro2021csdi, yan2021scoregrad}, audio processing~\cite{chen2020wavegrad, popov2021grad}, graph generation~\cite{huang2022graphgdp,niu2020permutation}, and bioinformatics \cite{xu2021geodiff,luo2022antigen}. Despite the significant interest and attention garnered by diffusion models, there remains a notable absence of an up-to-date and comprehensive taxonomy and analysis encapsulating the research advancements made in this field.

Diffusion models encompass two interconnected processes: a predefined forward process that maps the data distribution to a simpler prior distribution, often a Gaussian, and a corresponding reverse process that employs a trained neural network to gradually reverse the effects of the forward process by simulating Ordinary or Stochastic Differential Equations (ODE/SDE)~\cite{song2020score, karras2022elucidating}. The forward process resembles a straightforward Brownian motion with time-varying coefficients~\cite{karras2022elucidating}. The neural network is trained to estimate the score function utilizing the denoising score-matching objective~\cite{song2020denoising}. Consequently, diffusion models offer a more stable training objective compared to the adversarial objective employed in GANs and demonstrate superior generation quality when compared to VAEs, EBMs, and NFs~\cite{dhariwal2021diffusion, song2020score}.

However, it is imperative to acknowledge that diffusion models inherently entail a more time-intensive sampling process compared to GANs and VAEs. This stems from the iterative transformation of the prior distribution into a complex data distribution through the utilization of ODE/SDE (Ordinary/Stochastic Differential Equations) or Markov processes, necessitating a substantial number of function evaluations during the reverse process. Furthermore, additional challenges encompass the instability of the reverse process, the computational demands and constraints associated with training in high-dimensional Euclidean space, and the intricacies involved in likelihood optimization. In response to these challenges, researchers have put forth diverse solutions. For instance, advanced ODE/SDE solvers have been proposed to expedite the sampling process~\cite{lu2022dpm, zhang2022fast, Xu2023RestartSF}, while model distillation strategies have been employed~\cite{salimans2022progressive,Song2023ConsistencyM} to achieve the same goal. Furthermore, novel forward processes have been introduced to enhance sampling stability~\cite{Xu2022PoissonFG, Xu2023PFGMUT, dockhorn2021score} or facilitate dimensionality reduction~\cite{vahdat2021score, rombach2022high}. Additionally, a recent line of research endeavors to leverage diffusion models for efficiently bridging arbitrary distributions~\cite{Liu2022FlowSA, Albergo2022BuildingNF}. To provide a systematic overview, we categorize these advancements into four principal domains: \textbf{Sampling Acceleration}, \textbf{Diffusion Process Design}, \textbf{Likelihood Optimization}, and \textbf{Bridging Distributions}. Moreover, this survey will comprehensively examine the diverse applications of diffusion models across different domains, including computer vision, natural language processing, healthcare, and beyond. It will explore how diffusion models have been successfully applied to tasks such as \textbf{Image Synthesis}, \textbf{Video Generation}, \textbf{3D Generation}, \textbf{Medical Analysis}, \textbf{Text Generation}, \textbf{Speech Synthesis}, \textbf{Time Series Generation}, \textbf{Molecule Design}, and \textbf{Graph Generation} . By highlighting these applications, we aim to showcase the practical utility and transformative potential of diffusion models in real-world scenarios. 

The remaining sections are structured as follows: \Secref{sec: preliminaries} provides an overview of the fundamental formulations and theories of diffusion models. \Secref{sec: diffusion_algorithm} explores the algorithmic improvements made in the field, while \Secref{sec: diffusion_application} presents categorized applications based on the generation mechanism. Finally, in \Secref{sec: conclusion}, we summarize the content, discuss connections with other diffusion surveys, and identify limitations and future directions for diffusion models.

\section{Preliminaries}
\label{sec: preliminaries}

\subsection{Notions and Definitions}

\subsubsection{Time and States}

In diffusion models, the process unfolds over a timeline, which can be either continuous or discrete. The states within this timeline represent data distributions that describe the model's progression. Noise is incrementally added to the initial distribution, denoted as the starting state $x_0$, which is sampled from the data distribution $p_0$. The distribution gradually converges towards a known noise distribution, typically Gaussian, referred to as the prior state $x_T$. The states between the starting and prior states are intermediate states $x_t$, each associated with a marginal distribution $p_t$. This enables diffusion models to explore the evolution of the data distribution over time and generate samples that approximate the prior state $x_T$. The progression occurs through a sequence of intermediate states, with each state mapping to a specific time point in the diffusion process.

\subsubsection{Forward / Reverse Process, and Transition Kernel}
\label{sec: basicformulation}

In diffusion models, the forward process $F$ transforms the starting state into prior Gaussian noise, while the reverse process $R$ denoises the prior state back to the starting state using transition kernels. Following DDPM~\cite{ho2020denoising}, the discrete formulation for diffusion model is generalized by defining transition kernels among diffusion and denoising processes. 
\begin{equation}
F(x_0,\mathbf{\sigma}) = F_T(x_{T-1}, \sigma_T) \cdots \circ F_t(x_{t-1}, \sigma_t) \cdots \circ F_1(x_{0}, \sigma_1) \label{eq:forward}
\end{equation}
\begin{equation}
R(x_T,\mathbf{\sigma}) = R_1(x_{1}, \sigma_1) \cdots \circ R_t(x_{t}, \sigma_t) \cdots \circ R_T(x_T, \sigma_T) \label{eq:backward}
\end{equation}
where $F_t$ and $R_t$ denote the forward and reverse transition kernels at time $t$, with the noise scale $\sigma_t$ from noise set $\mathbf{\sigma}$. Unlike normalizing flow models, diffusion models incorporate variable noise, gradually refining the distribution for a controlled shift towards the target distribution, which provides a wider generation space and controllable generation. The discrete framework provides a discrete-time approximation of the continuous diffusion process, allowing for practical implementation and efficient computation.

\subsubsection{From discrete to continuous}

When the perturbation kernel is sufficiently small, the discrete processes (\Eqref{eq:forward} and \Eqref{eq:backward}) can be generalized to continuous processes. \cite{song2020score} showed that diffusion models with discrete Markov chains~\cite{sohl2015deep, ho2020denoising} can be incorporated into a continuous Stochastic Differential Equation (SDE) framework, where the generative process reverses a fixed forward diffusion process. A reserve ODE marginally equivalent to the reverse SDE has also been derived~\cite{song2020score}. The continuous process enjoys better theoretical support, and opens the door for applying existing techniques in the ODE/SDE community to diffusion models.

\subsection{Background}

In this sub-section, we introduce three foundation formulations \textbf{Denoised Diffusion Probabilistic Models}, \textbf{Score SDE Formulation}, and \textbf{Conditional Diffusion Probabilistic Models}, establishing the connection to \Secref{sec: diffusion_algorithm} and \Secref{sec: diffusion_application}. The following math formulation can be regarded as specific form of the general framework in \Secref{sec: basicformulation}.

\begin{figure*}[ht]
\centering
\includegraphics[width=1.00\textwidth]{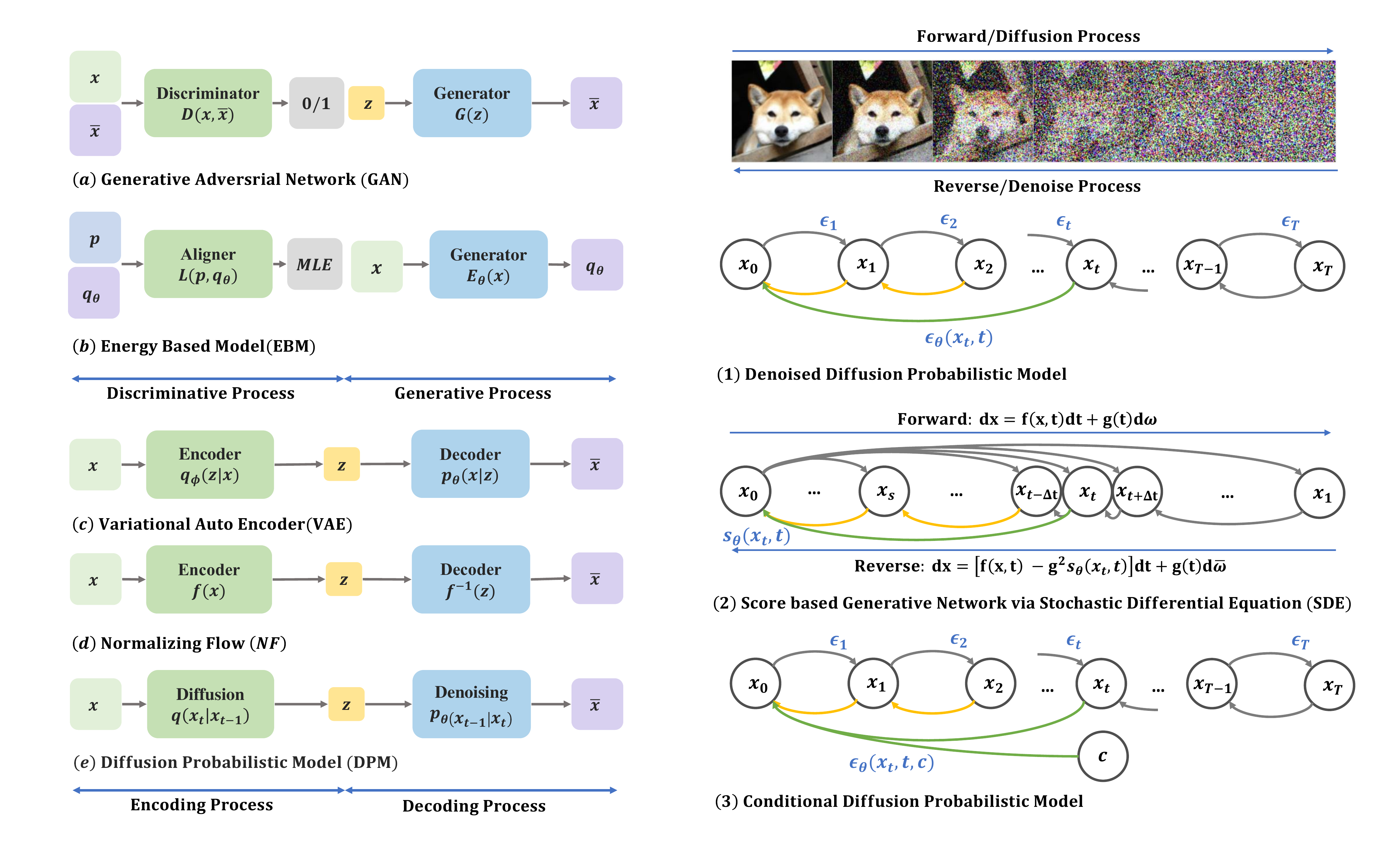}   
\caption{\textbf{Overview of Diffusion Models.} \textbf{(Left)} Briefing of Generative Models. GAN and EBM firstly utilize generators to capture data distribution. By sampling from training data, the estimated distributions are corrected based on real distributions. The distribution capturing processes as well as discrimination criteria differ. VAE, NF, and DPM directly project real distribution into pre-defined distributions by encoding process. Instances are obtained by decoding the samples from pre-defined distributions. They apply distinct pre-defined distributions $\mathbf{z}$ and encoding $\&$ decoding processes. \textbf{(Right)} Simplified Formulations of Diffusion Models. General procedures follow the top-right figure. Data distributions are diffused into random Gaussian noise, and are reversed by denoising. \textbf{(1) DDPM}(~\Secref{sec: ddpm}) achieves step-by-step diffusion and denoising processes along discrete timeline. \textbf{(2) SDE}(~\Secref{sec: sde}) establishes continuous timeline, achieving inter-state translation by function-based Stochastic Differential Equations (SDE). \textbf{(3) CDPM}(~\Secref{sec: cdpm}) employs condition $c$ in each sampling step of DPM to achieve controllable generation.}
\label{fig:Diffusion Model}
\vspace{-1mm}
\end{figure*}

\subsubsection{Denoised Diffusion Probabilistic Models (DDPM)}
\label{sec: ddpm}
 
\textbf{DDPM Forward Process:} In the DDPM framework, a sequence of noise coefficients $\beta_1, \beta_2, ..., \beta_T$ for Markov transition kernels are chosen, following patterns like constant, linear, or cosine schedules, leading to improved sample quality. According to \cite{ho2020denoising}, the forward steps are defined as:
\begin{equation}
F_t(x_{t-1}, \beta_t) := q(x_t|x_{t-1}) := \mathcal{N}\left(x_{t}; \sqrt{1-\beta_t}x_{t-1}, \beta_t\textbf{I}\right)
\end{equation}
By the composition of forward transition kernels from $x_0$ to $x_T$, the Forward Diffusion Process, which adds Gaussian noises to the data through the Markov kernel $q(x_t|x_{t-1})$:
\begin{equation}
F(x_0, \{\beta_i\}_{i=1}^T) := q\left(x_{1: T} \mid {x}_{0}\right) :=\prod_{t=1}^{T} q\left({x}_{t} \mid x_{t-1}\right)
\end{equation}
\textbf{DDPM Reverse Process:} The Reverse Process, with learnable Gaussian kernels parameterized by $\theta$, is defined as:
\begin{equation}
R_t(x_t, \Sigma_{\theta}) := p_{\theta}(x_{t-1}|x_t) := \mathcal{N}\left(x_{t-1}; \mu_{\theta}(x_t, t), \Sigma_{\theta}(x_t, t)\right)
\end{equation}
$\mu_{\theta}$ and $\Sigma_{\theta}$ are learnable mean and variance of the reverse Gaussian kernels, determined by reverse-step distribution $p_{\theta}$. The sequence of reverse steps from $x_T$ to $x_0$ is:
\begin{equation}
R(x_T, \Sigma_{\theta}) := p_{\theta}\left(x_{0: T}\right):=p\left({x}_{T}\right) \prod_{t=1}^{T} p_{\theta}\left({x}_{t-1} \mid {x}_{t}\right)
\end{equation}
DDPM aims to approximate the data distribution $p_0$ by the joint probability  distribution $p_{\theta}({x_0}) = \int p_{\theta}({x}_{0:T}) d{x}_{1:T}$.

\textbf{Diffusion Training Objective:} 
The training objective is equivalent to minimizing the variational bound on the negative log-likelihood by introducing KL-Divergence $D_{\mathrm{KL}}$:    
\begin{equation}
\begin{aligned}
\mathbb{E}\left[-\log p_{\theta}\left(\mathrm{x}_{0}\right)\right] &\leq \mathbb{E}_{q}[\underbrace{D_{\mathrm{KL}}\left(q\left({x}_{T} \mid {x}_{0}\right) \| p\left({x}_{T}\right)\right)}_{L_{T}} \\
& +\sum_{t>1} \underbrace{D_{\mathrm{KL}}\left(q\left({x}_{t-1} \mid {x}_{t}, {x}_{0}\right) \| p_{\theta}\left({x}_{t-1} \mid {x}_{t}\right)\right)}_{L_{t-1}} \\
& \underbrace{-\log p_{\theta}\left({x}_{0} \mid {x}_{1}\right)}_{L_{0}}] 
\end{aligned}
\end{equation}
where $L_T$ and $L_0$ denote the prior loss and the reconstruction loss; $L_{1:T-1}$ denoted the divergence sum between the posterior of the forward and reverse steps at the same time. Simplifying $L_{t-1}$, we obtain the simplified training objective named $L_{simple}$ based on the posterior $q({x}_{t-1}|{x}_t, {x}_0)$ as:
\begin{equation}
q\left({x}_{t-1} \mid {x}_{t}, {x}_{0}\right)=\mathcal{N}\left({x}_{t-1} ; \tilde{\boldsymbol{\mu}}_{t}\left({x}_{t}, {x}_{0}\right), \tilde{\beta}_{t} {I}\right)
\end{equation}
where $\tilde{\beta_t}$ depends on $\beta_t$. Keeping above parameterization and reparameterizing ${x}_t$ as $x_t(x_0, \sigma)$, $L_{t-1}$ is expressed as expectations of $\ell_2$-loss between two mean coefficients: 
\begin{equation}
L_{t-1}=\mathbb{E}_{q}\left[\frac{1}{2 \sigma_{t}^{2}}\left\|\tilde{\boldsymbol{\mu}}_{t}\left(\mathrm{x}_{t}, \mathrm{x}_{0}\right)-\boldsymbol{\mu}_{\theta}\left(\mathrm{x}_{t}, t\right)\right\|^{2}\right]+C
\end{equation}
which is linked to the denoising score-matching discussed in the next paragraph. Simplifying $L_{t-1}$ by reparameterizing $\mu_{\theta}$ w.r.t $\epsilon_{\theta}$, the simplified training objective named $L_{simple}$:
\begin{equation}
{L}_{simple} := \mathbb{E}_{{x}_0, {\epsilon}}
\left[\frac{{\beta_t}^2}{2{\sigma_T}^2\alpha_t(1-\bar{\alpha}_t)}\right]
\left\|{\epsilon}-\boldsymbol{\epsilon}_{\theta}(\sqrt{\bar{\alpha}_t}{x}_0 + \sqrt{1-\bar{\alpha}_t}{\epsilon})\right\|^{2} \label{eq:obj-ddpm}
\end{equation} 
Most diffusion models use the DDPM training strategy. However, Improved DDPM proposes combining $L{simple}$ with other objectives. After training, the prediction network $\boldsymbol{\epsilon}_{\theta}$ is used in the reverse process for ancestral sampling.

\subsubsection{Score SDE Formulation}
\label{sec: sde}

Score SDE~\cite{song2020score} extends the discrete-time scheme in DDPM to a continuous-time framework based on the stochastic differential equation. Also, it proposes additional deterministic sampling frameworks based on ODE formulation.

\textbf{Forward SDE:} ScoreSDE~\cite{song2020score} connected continuous diffusion process and stochastic differential equations. The reverse process is linked with the solution to Itô SDE~\cite{arnold1974stochastic} composed of a drift term for mean shift and a Brownian motion for additional noising:
\begin{equation}
d{x} = {f}(x,t)dt + g(t)d{w}, t \in [0, T] \label{eq:forward-sde}
\end{equation}
where ${w_t}$ is the standard Wiener process, ${f}(\cdot, t)$ is ${x}(t)$'s drift coefficient, and ${g}(\cdot)$ is a simplified diffusion coefficient independent on ${x}$. $p_t({x})$ and $p_T$ denote the marginal and prior distributions respectively. If coefficients are piece-wise continuous, the forward SDE equation has a unique solution~\cite{oksendal2013stochastic}. Two types of forward processes are proposed: Variation Preserving (VP) and Variation Explosion (VE) SDE. VP corresponds to the continuous extension of the DDPM framework:
\begin{align*}
    \textrm{VP:} &\quad \mathrm{d} {x}=-\frac{1}{2} \beta(t) {x ~ d} t+\sqrt{\beta(t)} \mathrm{d} w\\
    \textrm{VE:} &\quad \mathrm{d} {x}=\sqrt{\frac{\mathrm{d}\left[\sigma^{2}(t)\right]}{\mathrm{d} t}} \mathrm{~d} {w}
\end{align*}

\textbf{Reversed SDE:} The sampling of diffusion models is done via a corresponding reverse-time SDE of the forward process~(\Eqref{eq:forward-sde})~\cite{anderson1982reverse}:
\begin{equation}
\mathrm{d} {x}=\left[{f}({x}, t)-g(t)^{2} \nabla_{{x}} \log p_{t}({x})\right] \mathrm{d} \bar{t}+g(t) \mathrm{d} \overline{{w}}, t \in [0, T]\label{eq:reverse-sde}
\end{equation}
where ${w_t}$ is the standard Wiener process, ${f}(\cdot, t)$ is ${x}(t)$'s drift coefficient, and ${g}(\cdot)$ is a simplified diffusion coefficient. $p_t({x})$ and $p_T$ are the marginal and prior distributions. If coefficients are piece-wise continuous, a unique solution exists for the forward SDE equation.~\cite{vincent2011connection}: 
\begin{equation}
\begin{aligned}
{L} := \mathbb{E}_{t}&\{\lambda(t) \mathbb{E}_{{x}_0}  \mathbb{E}_{q({x}_t| {x}_0)}[\|{s}_{\boldsymbol{\theta}}({x}_t t)-\nabla_{{x}_t} \log p({x}_t| x_0)\|_{2}^{2}]\}   \label{eq:dsm}
\end{aligned}
\end{equation}
where $x_0$ is sampled from distribution $p_0$ and $\lambda(t)$ is the positive weighting function to keep the time-dependent loss at the same magnitude~\cite{song2020score}. $q({x}_t| {x}_0)$ is the Gaussian transition kernel associated with the forward process in \Eqref{eq:forward-sde}. For example, $q({x}_t| {x}_0)=\mathcal{N}\left(x_{t};x_0, \sigma^2(t)\textbf{I}\right)$. One can show that the optimal solution in the denoising score-matching objective~(\Eqref{eq:dsm}) equals the true score function  $\nabla_x\log p_t(x)$ for almost all $x, t$. Additionally, the score function $s_\theta$ can be seen as reparameterization of the neural prediction $\boldsymbol{\epsilon}_{\theta}$ in the DDPM objective~(\Eqref{eq:obj-ddpm}). \cite{Xu2023StableTF} further shows that the score function in the forward process of diffusion models can be decomposed into three phases. When moving from the near field to the far field, the perturbed data get influenced by more modes in the data distribution.

\textbf{Probability Flow ODE:} probability flow ODE~\cite{song2020score} supports the deterministic process which shares the same marginal probability density with SDE. Inspired by Maoutsa \textit{et al.} \cite{maoutsa2020interacting} and Chen \textit{et al.} \cite{chen2018neural}, any type of diffusion process can be derived into a special form of ODE. The corresponding probability flow ODE of \Eqref{eq:reverse-sde} is 
\begin{equation}
d{x} =  \{f({x}, t) - \frac{1}{2}{g(t)^2} \nabla_{{x}} \log p_t({x}) \}dt \label{eq:reverse-ode}
\end{equation}
In contrast to SDE, probability flow ODE can be solved with larger step sizes as they have no randomness. Thus, several works such as PNDMs \cite{liu2022pseudo} and DPM-Solver \cite{lu2022dpm1} obtain faster sampling speed based on advanced ODE solvers.

\subsection{Conditional Diffusion Probabilistic Models}
\label{sec: cdpm}

Diffusion models are versatile, capable of generating data samples from both unconditional $p_0$ and conditional $p_0(x|c)$ distributions, with $c$ as a given condition such as a class label or text linked to data $x$\cite{rombach2022high}. The score network $s_\theta(x, t, c)$ integrates this condition during training. Various sampling algorithms, including classifier-free guidance\cite{ho2022classifier} and classifier guidance~\cite{dhariwal2021diffusion}, are designed for conditional generation.

\textbf{Labeled Conditions} 
Sampling with labeled conditions guides each sampling step's gradient. It typically requires an additional classifier with a UNet Encoder architecture to generate condition gradients for specific labels, which can be text, categorical, binary, or extracted features~\cite{dhariwal2021diffusion,nichol2021glide,lu2022dpm,meng2022distillation,hu2022global,wolleb2022diffusion,packhauser2022generation,chen2022diffusiondet,baranchuk2021label,ho2022video}. The method, first presented by~\cite{dhariwal2021diffusion}, underpins current conditional sampling techniques.

\textbf{Unlabeled Conditions} 
Unlabeled condition sampling uses self-information for guidance, often applied in a self-supervised manner~\cite{hu2022self,chao2022quasi}. It is commonly used in denoising~\cite{chung2022come}, paint-to-image~\cite{choi2021ilvr}, and inpainting tasks~\cite{tashiro2021csdi}.

\begin{table*}
  \caption{\textbf{Taxonomy of Improvements on Diffusion Algorithms}(\Secref{sec: diffusion_algorithm})}
  \centering
\begin{forest}
  for tree={
  grow=east,
  reversed=true,
  anchor=base west,
  parent anchor=east,
  child anchor=west,
  base=left,
  rectangle,
  draw=black,
  rounded corners,align=left,
  minimum width=3em,
  edge+={darkgray, line width=1pt},
  inner xsep=4pt,
  inner ysep=1pt,
  },
  where level=1{text width=5em,fill=orange!10}{},
  where level=2{text width=5em,fill=blue!10}{},
  where level=3{yshift=0.26pt,fill=pink!30}{},
  where level=4{yshift=0.26pt,fill=yellow!20}{},
  where level=5{yshift=0.26pt}{},
  [\textbf{Diffusion} \\ \textbf{Algorithm} \\ \textbf{Improvement}, text width=8em, fill=green!20,
    [\textbf{Sampling} \\ \textbf{Acceleration}, text width=7em
        [Knowledge \\ Distillation, text width=5em,
            [ODE Trajectory, text width=6.6em
                [Progressive Distill \cite{salimans2022progressive}/TRACT~\cite{berthelot2023tract} \\ 
                Denoising Student \cite{luhman2021knowledge}/DSNO~\cite{zheng2022fast} \\
                Consistency Model~\cite{Song2023ConsistencyM}/RCFD~\cite{sun2022accelerating} \\
                Recfied Flow~\cite{liu2022flow}/SFT-PG~\cite{fan2023optimizing} \\
                MMD-DDM~\cite{aiello2023fast}/~\cite{lee2023minimizing,Meng2022OnDO}
                ]
            ]
            [SDE Trajectory, text width=6.6em
                [Recfied Flow~\cite{Liu2022FlowSA}/I2SB~\cite{Liu2023I2SBIS} \\ 
                Stochastic interpolant~\cite{Albergo2022BuildingNF}/DDIB~\cite{su2022dual}
                ]
            ]
        ]
        [Training \\ Scheme ,text width=4.5em
            [Diffusion Scheme \\ Learning ,text width=8em
                [TDPM\cite{zheng2022truncated}/Blurring Diffusion\cite{hoogeboom2022blurring} \\
                 ES-DDPM\cite{lyu2022accelerating}/Soft Diffusion\cite{daras2022soft} \\            
                 CCDF\cite{chung2022come}/\cite{https://doi.org/10.48550/arxiv.2206.05173,khrulkov2022understanding}
                ]
            ]
            [Noise Scale Design, text width=8.4em
                [VDM~\cite{kingma2021variational}/Improved DDPM~\cite{nichol2021improved} \\ FastDPM~\cite{kong2021fast}/\cite{san2021noise}
                ]
            ]
          ]
        [Training-Free \\ Sampling ,text width=6.6em
            [ODE, text width=3em                [DDIM~\cite{song2020denoising}/gDDIM~\cite{zhang2022gddim}/EDM~\cite{karras2022elucidating}  \\
                DEIS~\cite{zhang2022fast}/PNDM~\cite{liu2022pseudo}/DPM-Solver~\cite{lu2022dpm1}
            ]
            ]
            [SDE, text width=3em
                [Gotta Go Fast~\cite{zhang2022fast}  EDM~\cite{karras2022elucidating}/Restart~\cite{Xu2023RestartSF}
                ]
            ]       
          [Analytical, text width=4.4em
            [Analytic-DPM~\cite{bao2022analytic}/SN\&PNR-DDPM~\cite{bao2022estimating}
            ]
          ]
          [Dynamic \\ Programming, text width=6em
            [DDSS~\cite{watsonlearning}/Efficient Sampling\cite{watson2021learning}
            ]
          ]
        ]
        [Model \\ Merging, text width=4.5em
          [GAN-based, text width=5em
            [TDPM~\cite{zheng2022truncated}/Denoising GAN \cite{xiao2021tackling}
            ]
          ]
          [VAE-based, text width=5em
            [DiffuseVAE\cite{https://doi.org/10.48550/arxiv.2201.00308}/ES-DDPM\cite{lyu2022accelerating}
            ]
          ]
        ]
    ]
        [\textbf{Diffusion}\\ \textbf{Process Design}, text width=7em
            [Latent Space, text width=6em
              [LSGM~\cite{vahdat2021score}/INDM~\cite{kim2022maximum}/Latent Diffusion~\cite{rombach2022high}/DVDP~\cite{zhang2022dimensionality}
              ]
            ]
          [Innovative \\ Forward\\ Processes, text width=6em
              [PFGM~\cite{Xu2022PoissonFG}/PFGM++~\cite{Xu2023PFGMUT}/Cold Diffusion\cite{bansal2022cold}\\ 
              Flow-Matching~\cite{Lipman2022FlowMF}/EDM~\cite{karras2022elucidating}/CLD~\cite{dockhorn2021score}
              ]
          ]
          [Non-Euclidean, text width=6.5em
            [Discrete, text width=4em,
               [D3PM~\cite{austin2021structured}/Argmax~\cite{hoogeboom2021argmax}/ARDM~\cite{hoogeboom2021autoregressive}\\
              VQ-diffusion\cite{gu2022vector}/VQ-Diffusion+\cite{tang2022improved}/\cite{campbell2022continuous}
               ]
            ]
            [Manifold, text width=5em
              [RGSM~\cite{de2022riemannian}/PNDM~\cite{liu2022pseudo}/RDM~\cite{huang2022riemannian} \\
              Boomerang~\cite{luzi2022boomerang}/\cite{cheng2022theory}
              ]
            ]
            [Graphs, text width=5em
              [EDP-GNN~\cite{niu2020permutation}/Graph GDP~\cite{huang2022graphgdp} \\
              NVDiff \cite{chen2022nvdiff}/\cite{luo2022fast}
              ]
            ]
          ]
        ]
          [\textbf{Likelihood} \\ \textbf{Optimization}, text width=7em
            [MLE Training, text width=7.5em 
                [ScoreFlow~\cite{song2021maximum}/VDM~\cite{kingma2021variational}/\cite{huang2021variational}
                ]
            ]
            [Hybrid Loss, text width=7.5em
                [improved-DDPM~\cite{nichol2021glide}/\cite{Lu2022MaximumLT}
                ]
            ]
          ]
          [\textbf{Bridging} \\ \textbf{Distributions}, text width=7em
            [$\alpha$-blending~\cite{Heitz2023IterativeA}/Recfied Flow~\cite{Liu2022FlowSA}/I2SB~\cite{Liu2023I2SBIS}\\
            Stochastic interpolant~\cite{Albergo2022BuildingNF}/DDIB~\cite{su2022dual}, text width=22em
            ]
          ]
]
\end{forest}
\end{table*}

\section{Algorithm Improvement}
\label{sec: diffusion_algorithm}

Despite the high-quality generation of diffusion models across diverse data modalities, their real-world application could be improved. They necessitate a slow iterative sampling process, unlike other generative models like GANs and VAEs, and their forward process operates in high-dimensional pixel space. This section highlights four recent developments for enhancing diffusion models: 
\textbf{(1) Sampling Acceleration techniques}~(\Secref{sec:alg-sampling}) to speed up the standard ODE/SDE simulation; 
\textbf{(2) New Forward Processes}~(\Secref{sec:alg-forward}) for improved Brownian motion in pixel space; 
\textbf{(3) Likelihood Optimization techniques}~(\Secref{sec:alg-likelihood}) to enhance the diffusion ODE likelihood; 
\textbf{(4) Bridging Distribution techniques}~(\Secref{sec:alg-bridge}) that utilize diffusion model concepts to connect two distinct distributions.

\subsection{Sampling Acceleration}
\label{sec:alg-sampling}
Despite their high-fidelity generation, the practical utility of diffusion models is limited by their slow sampling speed. This section briefly overviews four advanced techniques to enhance sampling speed: distillations, training schedule optimization, training-free acceleration, and integration of diffusion models with faster generative models.

\subsubsection{{Knowledge Distillation}}

Knowledge distillation, a technique for transferring "knowledge" from larger to simpler models, is becoming increasingly popular~\cite{lopes2017data,gou2021knowledge}. In diffusion models, the goal is to produce samples using fewer steps or smaller networks by aligning and minimizing the discrepancy between original and generated samples. Viewed as trajectory optimization across distributions, distillation offers optimal mappings for cost-effective and faster controllable generation.

\textbf{ODE Trajectory}
Knowledge distillation from teacher to student models using ODE formulation parallels mapping prior distribution to target distribution via efficient paths across the distribution field. \cite{salimans2022progressive} first applied this principle to improve diffusion models by progressively distilling sampling trajectories, straightening latent mappings every two steps. TRACT~\cite{berthelot2023tract}, Denoising Student~\cite{luhman2021knowledge}, and Consistency Models~\cite{Song2023ConsistencyM} extended this effect, increasing acceleration rates to 64 and 1024, by directly estimating clean data from noisy samples at time $T$. RFCD~\cite{sun2022accelerating} enhances student model performance by aligning sample features during training.

Optimal trajectories can be obtained through optimal transport~\cite{villani2003topics}. By minimizing transportation cost among distributions via flow matching, ReFlow~\cite{liu2022flow} and \cite{lee2023minimizing} achieve one-step generation. DSNO~\cite{zheng2022fast} proposes a neural operator for direct temporal path modeling. Consistency Model~\cite{Song2023ConsistencyM}, SFT-PG~\cite{fan2023optimizing}, and MMD-DDM~\cite{aiello2023fast} search ideal trajectories using LPIPS, IPA, and MMD, respectively.

\textbf{SDE Trajectory}
Distilling stochastic trajectories is still challenging. Few works are proposed (referred to Section~\ref{sec:alg-bridge}).

\subsubsection{Training Schedule}

Improving the training schedule involves modifying traditional training settings, such as diffusion schemes and noise schemes, that are independent of sampling. Recent research has highlighted the crucial factors in training schemes that impact learning patterns and model performance. In this subsection, we categorize training enhancements into two main areas: diffusion scheme learning and noise scale design.

\textbf{Diffusion Scheme Learning}
Diffusion models, which project data into latent spaces like Variational Autoencoders (VAEs), are more complex due to their higher expressiveness. Reverse decoding methods in these models can be divided into two approaches: encoding degree optimization and projecting approaches.

Encoding degree optimization methods, such as CCDF \cite{chung2022come} and Franzese et al. \cite{https://doi.org/10.48550/arxiv.2206.05173}, minimize the Evidence Lower Bound (ELBO) by treating the number of diffusion steps as a variable. Truncation, another approach, balances generation speed and sample fidelity by sampling from less diffused data in a one-step manner. TDPM \cite{zheng2022truncated} and ES DDPM \cite{lyu2022accelerating} use truncation with GAN and CT \cite{zheng2020act}.
Projecting approaches, like Soft diffusion \cite{daras2022soft} and blurring diffusion models \cite{hoogeboom2022blurring}, explore the diversity of diffusion kernels using linear corruptions such as blurring and masks.

\textbf{Noise Scale Designing}
In traditional diffusion processes, each transition step is determined by injected noise, which is equivalent to a random walk on forward and reversed trajectories. Designing the noise scale can lead to reasonable generation and fast convergence. Unlike traditional DDPMs, existing methods treat the noise scale as a learnable parameter throughout the process.

Forward noise design methods like VDM \cite{kingma2021variational} parameterize the noise scale as a signal-to-noise ratio, connecting it to training loss and model types. FastDPM \cite{kong2021fast} links noise design to ELBO optimization using discrete-time variables or a variance scalar. For reverse noise design, improved DDPM \cite{nichol2021improved} learns the reverse noise scale implicitly by training a hybrid loss, while San Roman \textit{et al.} use a noise prediction network to update the reverse noise scale before ancestral sampling.

\subsubsection{Training-Free Sampling}

Training-free methods aim to leverage advanced samplers to accelerate the sampling process of pre-trained diffusion models, eliminating the need for model re-training. This subsection categorizes these methods into several aspects: acceleration of the diffusion ODE and SDE samplers, analytical methods, and dynamic programming.

\textbf{ODE Acceleration}
\cite{song2020score} demonstrates that the stochastic sampling process in DDPM has a marginally-equivalent probability ODE, which defines deterministic sampling trajectories from prior to data distribution. Given that ODE samplers generate less discretization error than their stochastic counterparts~\cite{song2020score, Xu2023RestartSF}, most previous work on sampling acceleration has been ODE-centric. For instance, the widely-used sampler DDIM~\cite{song2020denoising} can be regarded as a probability flow ODE~\cite{song2020score}:
\begin{equation}
\mathrm{d} \overline{{x}}(t)=\epsilon_{\theta}^{(t)}\left(\frac{\overline{{x}}(t)}{\sqrt{\sigma^{2}+1}}\right) \mathrm{d} \sigma(t)
\end{equation}
where $\sigma_t$ is parameterized by $\sqrt{1-\alpha_t}/\sqrt{\alpha_t}$, and $\bar{x}$ is parameterized as ${x}/\sqrt{\alpha_t}$. Later works~\cite{zhang2022fast, lu2022dpm1} interpret DDIM as a product of applying an exponential integrator on the ODE of Variance Preserving~(VP) diffusion~\cite{song2020score}. Advanced ODE solvers have been utilized in methods such as PNDM \cite{liu2022pseudo}, EDM \cite{karras2022elucidating}, DEIS \cite{zhang2022fast}, gDDIM \cite{zhang2022gddim}, and DPM-Solver \cite{lu2022dpm1}. For example, EDM employs Heun's $2^{\textrm{nd}}$ order ODE solvers, and DEIS/DPM-solver improves upon DDIM by numerically approximating the score functions within each discretized time interval. These methods significantly accelerate the sampling speed (reducing the number of function evaluations, or NFE) compared to the original DDPM sampler while still yielding high-quality samples.

\textbf{SDE Acceleration}
ODE-based samplers are faster but reach performance limits, while SDE-based samplers offer better sample quality despite being slower. Several works have focused on accelerating stochastic samplers' speed. Gotta Go Fast~\cite{jolicoeur2021gotta} uses adaptive step size for faster SDE sampling, while EDM~\cite{karras2022elucidating} combines higher-order ODE with Langevin-dynamics-like noise addition and removal, demonstrating that their proposed stochastic sampler significantly outperforms the ODE sampler on ImageNet-64. A recent work~\cite{Xu2023RestartSF} reveals that although ODE-samplers involve smaller discretization errors, the stochasticity in SDE helps to contract accumulated errors. This leads to the Restart Sampling algorithm~\cite{Xu2023RestartSF}, which blends the best aspects of both worlds. The sampling method alternates between adding significant noise by additional forward steps and strictly following a backward ODE, surpassing previous SDE and ODE samplers on standard benchmarks and the Stable Diffusion model~\cite{rombach2022high}, both in terms of speed and accuracy.

\textbf{Analytical Method}
Existing training-free sampling methods treat reverse covariance scales as a hand-crafted sequence of noises without considering them dynamically. Starting from KL-divergence optimization, analytical methods set the reverse mean and covariance using the Monte Carlo method. Analytic-DPM \cite{bao2022analytic} and extended Analytic-DPM \cite{bao2022estimating} jointly propose optimal reverse solutions under correction for each state. Analytical methods enjoy a theoretical guarantee for the approximation error, but they are limited to specific distributions due to their pre-assumptions.

\textbf{Dynamic Programming Adjustment}
Dynamic programming (DP) achieves the traversal of all choices to find the optimized solution in a reduced time by using a memorization technique \cite{bellman1966dynamic}. Assuming that each path from one state to another state shares the same KL divergence with others, dynamic programming algorithms explore the optimal traversal along the trajectory. Current DP-based methods \cite{watson2021learning,watson2021learning22} take $\mathcal{O}\left(T^{2}\right)$ of computational cost by optimizing the sum of ELBO losses.

\subsubsection{Merging Diffusion and Other Generative Models}
Diffusion models can be synergized with other generative models like Generative Adversarial Networks (GANs) or Variational Autoencoders (VAEs) to streamline the sampling process. For example, pristine data $x_{0}$ can be directly predicted through a VAE~\cite{https://doi.org/10.48550/arxiv.2201.00308} or GAN \cite{xiao2021tackling} obtained from noisy samples during an intermediate phase of the diffusion sampling process. Moreover, a VAE~\cite{lyu2022accelerating} or GAN~\cite{zheng2022truncated} can generate samples at intermediary diffusion time steps, which are then denoised by diffusion models until time $t=0$ for faster time traversal.

\subsection{Diffusion Process Design}
\label{sec:alg-forward}
The traditional forward process in diffusion models, often considered as Brownian motion in pixel space~\cite{ho2020denoising, karras2022elucidating}, may be sub-optimal for generative modeling. Consequently, research efforts have been directed towards creating new diffusion processes that simplify and enhance the associated backward processes for neural networks. This path has bifurcated into developing latent spaces designed for diffusion models~(\Secref{sec:new-latent}) and replacing the conventional forward process with improved versions in pixel space~(\Secref{sec:new-forward}). Special attention is also given to diffusion processes specifically tailored for non-Euclidean spaces like manifolds, discrete spaces, functional spaces, and graphs~(\Secref{sec:new-space}).

\subsubsection{Latent Space}
\label{sec:new-latent}
Researchers explore training diffusion models in a learned latent space to enhance neural networks and establish a more direct backward process. This approach is exemplified by LSGM \cite{vahdat2021score} and INDM \cite{kim2021maximum}, which jointly train a diffusion model and a VAE or normalizing flow model. Both models share a common objective, the weighted denoising score-matching loss~($L_{DSM}$ in \Eqref{eq:dsm}), to optimize the pair of encoder-decoder and diffusion model.
\begin{equation}
    L := L_{Enc}(z_0|x)+L_{Dec}(x|z_0)+L_{DSM}\left(\left(\left\{z_{t}\right\}_{t=0}^{T}\right)\right) \label{eq:joint}
\end{equation}
Here, $z_0$ represents the latent form of the original data $x$, while $z_t$ is its perturbed counterpart. It is important to note that $z_t$ is a function of the encoder, hence the $L_{DSM}$ loss also updates the encoder's parameters. The joint objective is optimizing the ELBO or log-likelihood~\cite{vahdat2021score, kim2021maximum}. This leads to a latent space that is simpler to learn from and to sample. Influential work such as Stable Diffusion \cite{rombach2022high} separates the process into two stages: learning the latent space of VAE and training diffusion models with text as conditional inputs. On a different note, DVDP \cite{zhang2022dimensionality} decomposes the pixel space into orthogonal components and dynamically adjusts the attenuation of each component during image perturbation, akin to dynamic image down-sampling and up-sampling.

\subsubsection{Emerging Forward Processes}
\label{sec:new-forward}

Latent space diffusion has advantages but also adds complexity and computational load to the framework. To address this issue, contemporary research explores forward process design for more robust and rddicient generative models. For instance, the Poisson Field Generative Model (PFGM)\cite{Xu2022PoissonFG} treats data as electric charges in an augmented space, guiding a simple distribution along electric field lines towards the data distribution. The forward process in this model is defined in the electric field lines' directions, exhibiting more robust backward sampling than diffusion models. The PFGM++~\cite{Xu2023PFGMUT} extends PFGM with higher-dimensional augmented variables, and an interpolation between these models reveals an optimal point, leading to state-of-the-art image generation. PFGM and PFGM++ also find applications in antibody~\cite{huang2023pf} and medical image~\cite{Ge2023JCCSPFGMAN} generation.

Dockhorn et al. \cite{dockhorn2021score} introduced the Critically-Damped Langevin Diffusion (CLD) model, which incorporates "velocity" variables interacting through Hamiltonian dynamics. The model simplifies learning the score function of the conditional velocity distribution, compared to directly learning the data's score functions. Given the success of physics-inspired generative models such as diffusion models and PFGM, a recent work~\cite{Liu2023GenPhysFP} provides a systematic method to transform physical processes into generative models.

Other research explores alternative corrupting processes. For instance, Cold Diffusion\cite{bansal2022cold} uses arbitrary image transformations like blurring for the forward process, while \cite{Rissanen2022GenerativeMW} applies heat dissipation in pixel space. Furthermore, there are efforts to enhance training and sampling with advanced Gaussian perturbation kernels~\cite{karras2022elucidating, Lipman2022FlowMF}.

\subsubsection{Diffusion Models on non-Euclidean space}
\label{sec:new-space}

\textbf{Discrete Space}
Deep generative models have made considerable strides in various domains, such as natural language processing \cite{vaswani2017attention,devlin2018bert}, multi-modal learning \cite{nichol2021glide,ramesh2022hierarchical}, and AI for science \cite{jumper2021highly,ovchinnikov2021structure}. A key achievement is the processing of discrete data, including sentences, residues, atoms, and vector-quantized data. Diffusion models are commonly used in these applications, focusing on text, categorical data, and vector-quantized data. D3PM \cite{austin2021structured} defines the forward process in discrete space, processing data like text or atom type, using transition kernels $Q_t$: 
\begin{equation}
q\left({x}_{t} \mid {x}_{t-1}\right)=\operatorname{Cat}\left({x}_{t} ; {p}={x}_{t-1} \boldsymbol{Q}_{t}\right)
\end{equation}
where $\operatorname{Cat}()$ denotes a categorical distribution. This approach has been extended for generating language text, segmentation maps, and lossless compression \cite{hoogeboom2021argmax, hoogeboom2021autoregressive}. 

For multi-modal problems such as text-to-image generation and text-to-3d generation, vector-quantized (VQ) data transforms data into codes, achieving excellent performance in autoregressive encoders \cite{van2017neural}. Diffusion techniques were first applied to VQ data by \cite{gu2022vector}, addressing the unidirectional bias and accumulation prediction error in VQ-VAE. This core idea has been utilized in further text-to-image, text-to-pose, and text-to-multimodal works \cite{cohen2022diffusion, tang2022improved, xie2022vector, guo2022generating, weinbach2022m, xu2022versatile}. The forward process is defined by the probability transition matrix $Q$ and categorical representation vector $v$:
\begin{equation}
q\left(x_{t} \mid x_{t-1}\right)=\boldsymbol{v}^{\top}\left(x_{t}\right) \boldsymbol{Q}_{t} \boldsymbol{v}\left(x_{t-1}\right)
\end{equation}

\textbf{Manifold}
Data structures like images and videos typically inhabit Euclidean space. However, certain data in fields like robotics \cite{pierson2017deep}, geoscience \cite{de2019progress}, and protein modeling \cite{wang2018computational} are defined within a Riemannian manifold \cite{cao2020comprehensive}. Standard Euclidean methods may not apply in this environment. To address this, recent methodologies such as RDM \cite{huang2022riemannian}, RGSM \cite{de2022riemannian}, and Boomerang \cite{luzi2022boomerang} have incorporated diffusion sampling into the Riemannian manifold, extending the score SDE framework \cite{song2020score}. Theoretical works \cite{cheng2022theory,liu2022pseudo} provide further support for manifold sampling.

\textbf{Graph}
Graph-based neural networks are gaining popularity due to their expressiveness in human pose \cite{guo2022generating}, molecules \cite{lin2022diffbp}, and proteins \cite{anand2022protein} \cite{wu2021self}. Current methods apply diffusion theories to graphs. Approaches like EDP-GNN \cite{niu2020permutation}, Pan \textit{et al.} \cite{luo2022fast}, and GraphGDP \cite{huang2022graphgdp} process graph data via adjacency matrices to capture permutation invariance. NVDiff \cite{chen2022nvdiff} reconstructs node positions using reverse SDE.

\textbf{Function}
Dutordoir \textit{et al.}, \cite{dutordoir2022neural} introduced the first diffusion model sampling in functional space, capturing infinite-dimensional distributions via joint posterior sampling.

\subsection{Likelihood Optimization}
\label{sec:alg-likelihood}
While diffusion models~\cite{ho2020denoising} optimize the ELBO to overcome the intractability of the log-likelihood, the likelihood optimization is ignored, which is challenging for continuous-time diffusion models~\cite{song2020score}. Two approaches including MLE Training~(\Secref{sec:mle-training}) and hybrid loss~(\Secref{sec:hybrid-loss}) are designed to enhance likelihood training.

\subsubsection{MLE Training} 
\label{sec:mle-training}
Three concurrent works—ScoreFlow \cite{song2021maximum}, VDM~\cite{kingma2021variational}, and \cite{huang2021variational} establish a connection between the MLE training and the weighted denoising score-matching (DSM) objective in diffusion models, primarily through the use of the Girsanov theorem. For instance, ScoreFlow~\cite{song2021maximum} demonstrates that under a particular weighting scheme, the DSM objective provides an upper bound on the negative log-likelihood. This finding enables a neural-network parameter-independent approximation of score-based MLE.

\subsubsection{Hybrid Loss} 
\label{sec:hybrid-loss}
Instead of solely relying on maximum likelihood training, certain approaches introduce hybrid loss designs to improve the model likelihood in DSM. One such approach is Improved DDPM~\cite{nichol2021improved}, which proposes learning the variances of the reverse process using a simple reparameterization technique and a hybrid learning objective that combines the variational lower bound and DSM. Additionally, \cite{Lu2022MaximumLT} demonstrates that incorporating high-order score-matching loss contributes to enhancing the log-likelihood.

\subsection{Bridging Distributions} 
\label{sec:alg-bridge}

Diffusion models excel at transforming simple Gaussian distributions but face challenges when bridging arbitrary distributions, particularly in areas like image-to-image translation and cell distribution transportation. Various approaches have been proposed to tackle this issue.
One approach, known as $\alpha$-blending~\cite{Heitz2023IterativeA}, involves iterative blending and deblending to create a deterministic bridge. Diffusion models are treated as special cases when one end distribution is Gaussian. Another approach is Rectified Flow~\cite{Liu2022FlowSA}, which incorporates additional steps to straighten the bridge.
Other methods, such as the one proposed in \cite{Albergo2022BuildingNF}, suggest constructing an ODE with general interpolant functions between two distributions. Besides, others explore the utilization of the Schrödinger Bridge~\cite{Liu2023I2SBIS} or Gaussian distributions as junctions to connect two diffusion ODEs~\cite{su2022dual}.

\begin{table*}
  \caption{\textbf{Classification of Diffusion-based model Applications}(\Secref{sec: diffusion_application})}
  \centering
\begin{forest}
  for tree={
  grow=east,
  reversed=true,
  anchor=base west,
  parent anchor=east,
  child anchor=west,
  base=left,
  rectangle,
  draw=black,
  rounded corners,align=left,
  minimum width=2.5em,
inner xsep=4pt,
inner ysep=1pt,
  },
  where level=1{text width=5em,fill=blue!6}{},
  where level=2{text width=5em,fill=pink!30}{},
  where level=3{yshift=0.26pt,fill=yellow!20}{},
  where level=4{yshift=0.26pt}{},
  where level=5{yshift=0.26pt}{},
  [\textbf{Diffusion} \\ \textbf{Application}, fill=orange!10,
    [\textbf{Image} \\ \textbf{Generation}, text width=5em
        [Unconditional \& \\ Class Condition, text width=7.5em
            [DDPM\cite{ho2020denoising}/Imagen~\cite{saharia2022photorealistic}/diffuison beats gan~\cite{dhariwal2021diffusion}
            ]
        ]
        [Text \\ Condition, text width=5em
            [Imagen\cite{saharia2022photorealistic}/Stable Diffusion~\cite{rombach2022high}/DALL-E 2~\cite{ramesh2022hierarchical}/~\cite{hertz2022prompt,ruiz2023dreambooth,chefer2023attend}
            ]
        ]
        [Image \\ Condition, text width=5em
            [Instructpix2pix\cite{brooks2023instructpix2pix}/\cite{zhang2023adding}                
            ]
        ]
    ]
    [\textbf{3D} \\ \textbf{Generation}, text width=5em
        [3D Condition, text width=6em
            [PDR\cite{lyu2021conditional}/Shape-E~\cite{jun2023shap}/Point-E~\cite{nichol2022point}/PVD\cite{zhou20213d} \\
            Zero-1-to-3~\cite{Liu_2023_ICCV}/One-2-3-45~\cite{liu2023one}/\cite{luo2021score,luo2021diffusion}]
        ]
        [2D Condition, text width=6em
            [DreamFusion\cite{poole2022dreamfusion}/Magic3d\cite{lin2023magic3d}]
        ]
    ]
    [\textbf{Video} \\ \textbf{Generation}, text width=5em
        [Generation, text width=5em
            [VDM\cite{ho2022video}/Make-A-Video\cite{singer2022make}/MCVD\cite{voleti2022mcvd}/FDM\cite{harvey2022flexible} \\
            RVD\cite{yang2022diffusion}/RaMViD\cite{hoppe2022diffusion}/AnimatedDiff~\cite{guo2023animatediff}
            ]
        ]
    ]
    [\textbf{Medical} \\ \textbf{Analysis}, text width=5em
        [In-distribution, text width=6em
            [MCG~\cite{chung2022improving}/Score-MRI~\cite{chung2022score}/Diff-MIC~\cite{yang2023diffmic} \\
            OCT-DDPM~\cite{hu2022unsupervised}/CCDF\cite{chung2022come}]
        ]
        [Cross- \\ distribution, text width=6em
            [AnoDDPM~\cite{wyatt2022anoddpm}/FNDM~\cite{li2023fast}/DifuseMorph~\cite{kim2022diffusemorph} \\
            R2D2+~\cite{chung2022mr}/3D-DDPM-Med~\cite{dorjsembe2022three}/~\cite{gong2023diffusion,pinaya2022brain}
            ]
        ]
    ]
    [\textbf{Text} \\ \textbf{Generation}, text width=5em
        [Discrete
            [D3PM\cite{austin2021structured}/Argmax\cite{hoogeboom2021argmax}/DiffusionBERT\cite{he2022diffusionbert}          
            ]
        ]
        [Latent
            [Diffusion-LM\cite{li2022diffusion}/Seqdiffuseq~\cite{yuan2022seqdiffuseq}/GENIE~\cite{lin2023text}/LIVE\cite{tang2023learning} \\
             DiffuSeq~\cite{gong2022diffuseq}/AR-Diffusion~\cite{wu2023ar}/Difformer~\cite{gao2022difformer}/SED~\cite{strudel2022self}/~\cite{ye2023diffusion}
             ]
        ]
    ]
    [\textbf{Time Series} \\ \textbf{Generation}, text width=5.4em
        [Imputation, text width=5em
            [TSGM~\cite{lim2023regular}/CSDI\cite{tashiro2021csdi}/PriSTI~\cite{liu2023pristi}/SSSD~\cite{lopez2023diffusion}/TransFusion\cite{fahim2023transfusion}
            ]
        ]
        [Prediction, text width=5em
            [TimeGrad\cite{rasul2021autoregressive}/ScoreGrad~\cite{yan2021scoregrad}/DiffSTG~\cite{wen2023diffstg}
            ]
        ]
    ]
    [\textbf{Audio} \& \\ \textbf{Speech} \\ \textbf{Generation}, text width=5em
        [Conversion \& \\ Separation, text width=6em
            [WaveGrad\cite{chen2020wavegrad}/DiffWave\cite{kong2020diffwave}/DiffSinger\cite{liu2022diffsinger} \\
            ProDiff\cite{huang2022prodiff}/BinauralGrad\cite{leng2022binauralgrad}/DiffSVC\cite{liu2021diffsvc}/\cite{wu2021itotts}
            ]            
        ]
        [Content \\ Condition, text width=6em
            [EdiTTS\cite{tae2021editts}/Diff-TTS\cite{jeong21_interspeech}/SpecGrad\cite{koizumi2022specgrad}/~\cite{kim2022guided2} \\
            Guided-TTS\cite{kim2022guided}/DiffSound~\cite{yang2022diffsound}/DiffSinger~\cite{liu2022diffsinger}
            ]
        ]
    ]
    [\textbf{Molecule} \\ \textbf{Generation}, text width=5em
        [Unconditional, text width=6.4em             
            [GeoDiff\cite{xu2021geodiff}/EDM\cite{hoogeboom2022equivariant}/ProteinSGM\cite{lee2022proteinsgm} \\
            Torsional~\cite{jing2022torsional}/SE3Diffusion~\cite{yim2023se}/FoldingDIff~\cite{wu2022protein} \\
            ]
        ]
        [Multi-modal, text width=6.4em
            [DiffDock~\cite{corso2022diffdock}/DiffAb~\cite{luo2022antigen}/Co-Design~\cite{anand2022protein} \\ RFDiffusion~\cite{watson2023novo}/ProteinGenerator~\cite{lisanza2023joint}
            ]
        ]
    ]
    [\textbf{Graph} \\ \textbf{Generation}, text width=5em
        [Unconditional, text width=6.4em
            [GraphGDP~\cite{huang2022graphgdp}/DiGress~\cite{vignac2022digress}/EDP-GNN~\cite{niu2020permutation}]
        ] 
        [Conditional, text width=6.4em
            [PCFI~\cite{um2022confidence}/EDGE~\cite{chen2023efficient}/DiffFormer~\cite{wu2022difformer}/D4Explainer~\cite{chen2023d4explainer}
            ]
        ]
    ]
]
\end{forest}
\end{table*}

\section{Application}
\label{sec: diffusion_application}

Benefiting from the powerful ability to generate realistic samples, diffusion models have been widely used in various fields. In real-world applications, the key to unlocking the power of diffusion models lies in fitting the diffusion process, denoising process, and conditional sampling to the natural of a wide range of data. Inspired by this idea, the applications of diffusion are summarised as \textbf{Image Generation}, \textbf{3D Generation}, \textbf{Video Generation}, \textbf{Medical Analysis}, \textbf{Text Generation}, \textbf{Time Series Generation}, \textbf{Audio Generation}, \textbf{Molecule Design}, and \textbf{Graph Generation}.

\subsection{Image Generation}

Diffusion models have achieved remarkable performance on image generation, either on traditional class-conditioned or unconditional generation~\cite{dhariwal2021diffusion, ho2020denoising, saharia2022photorealistic}, or on more complicated text or image condition~\cite{rombach2022high, zhang2023adding}, or their combinations~\cite{brooks2023instructpix2pix}. Our discussion henceforth will concentrate on application settings that mimic real-world scenarios, categorizing applications according to the conditional inputs.

\subsubsection{Text condition}

Diffusion models demonstrate exceptional performance in text-to-image generation, capable of creating not only photorealistic images but also samples that closely adhere to user-provided textual inputs. Remarkable examples include Imagen~\cite{saharia2022photorealistic}, Stable Diffusion~\cite{rombach2022high} and DALL-E 2~\cite{ramesh2022hierarchical}. Built on top of existing diffusion architectures, these methods add a cross-attention layer to inject the sequence of text embeddings into the diffusion models. The experimental results show that such conditioning mechanism effectively blends the text information into the generated images. 

In addition, the cross-attention conditional mechanism enables many training-free image editing by utilizing and manipulating the keys, values, or attention matrices in the cross-attention layers. For example, \cite{hertz2022prompt} changes the concepts in source images by swapping or adding new feature maps into the output of the cross-attention layers; \cite{ruiz2023dreambooth} enables the customization of a new concept by learning a new text embedding as the input to the cross-attention layers. \cite{chefer2023attend} enforces the cross-attention to attend to all subject tokens in the text prompt and enlarge their activations, encouraging the model to faithfully generate all subjects described in the text prompt.

\subsubsection{Image condition}

In addition to textual conditions, diffusion models also support image conditions, such as images to be edited, depth maps, or human skeletons, as conditional inputs. The underlying concept remains the same, which involves incorporating encoded image features into the diffusion backbone. The work by \cite{brooks2023instructpix2pix} introduces encoded features from the source image into the first convolutional layer to enable image conditioning, thereby allowing for image-to-image editing with text prompts. Similarly, \cite{zhang2023adding} utilizes depth maps, Canny edges, or human skeletons to control the spatial layout of the generated images.

\subsection{3D Generation}

Broadly, there are two primary approaches to 3D generation by diffusion models. The first approach focuses on training these models directly with 3D data. However, due to the limited availability of 3D data, the second approach emphasizes generating 3D content by 2D diffusion priors.

\subsubsection{3D Data Condition}

Given the diverse range of 3D representations, such as NeRF, point clouds, voxels, Gaussian splatting, and more, diffusion models have been effectively applied across these various 3D representations. For instance, works such as \cite{luo2021diffusion, zhou20213d, luo2021score} directly generate point clouds for 3D objects. In order to achieve efficient sampling, a hybrid point-voxel representation was employed for shape processing in PDR~\cite{lyu2021conditional}, introducing a new paradigm for point cloud completion. Building upon this research, Point-E~\cite{nichol2022point} further incorporates image synthesis as an additional conditional input for point cloud diffusion models.

In contrast, Shape-E~\cite{jun2023shap} utilizes diffusion models for the NeRF representation of 3D objects. Zero-1-to-3~\cite{Liu_2023_ICCV} takes a different approach by training viewpoint-conditioned diffusion models to enable novel view synthesis. It then optimizes a NeRF based on the generated samples from different camera viewpoints. Based on this work, \cite{liu2023one} further extends Zero-1-to-3 by incorporating a pose estimation stage.

\subsubsection{2D Diffusion prior}

Another interesting line of works is aiming to \textit{distill 3D from a 2D diffusion model.} Dreamfusion~\cite{poole2022dreamfusion} smartly use the score distillation sampling~(SDS) objective to distill a NeRF from a pre-trained text-to-image models. They optimize a randomly initialized NeRF via gradient descent such that the rendered images from different angles achieve low loss. \cite{lin2023magic3d} extends DreamFusion to a two-stage coarse-to-fine optimization framework, to accelerate the generation process.

\subsection{Video Generation}

Video diffusion models augment the 2D diffusion models for image generation with an additional time axis. The general idea is to add a temporal layer to explicitly model the cross-frame dependence in existing 2D diffusion structures. Representative works include Video Diffusion Models~\cite{ho2022video}, Make-A-Video~\cite{singer2022make}, AnimatedDiff~\cite{guo2023animatediff}, RVD~\cite{yang2022diffusion}, FDM~\cite{harvey2022flexible}, MCVD~\cite{voleti2022mcvd}. RaMViD~\cite{hoppe2022diffusion} extends image diffusion models to videos with 3D convolutional neural networks and designed a conditioning technique for video prediction, infilling, and up-sampling.

\subsection{Medical Analysis}
Diffusion models provide a solution to the challenges encountered in medical analysis, where acquiring large-scale, high-quality annotated datasets is challenging. These models demonstrate exceptional performance in tasks related to in-distribution analysis and cross-distribution generation.

\subsubsection{In-distribution Analysis}
Diffusion models are effective in various medical imaging tasks, leveraging their ability to accurately capture medical images with strong prior information. They have been successfully used in super-resolution~\cite{chung2022improving,chung2022come}, classification~\cite{yang2023diffmic}, and noise robustness~\cite{chung2022score,hu2022unsupervised}. For example, Score-MRI~\cite{chung2022score} accelerates MRI reconstruction using pixel guidance SDE sampling, while Diff-MIC~\cite{yang2023diffmic} achieves accurate classification across multiple modalities with Dual-granularity guidance and Maximum-Mean Discrepancy. Additionally, MCG~\cite{chung2022improving} proposes manifold correction during sampling for CT super-resolution, reducing errors and improving acceleration.

\subsubsection{Cross-distribution Generation}
Multimodal guidance has significantly improved generative capabilities in medical analysis. By integrating class-specific guidance~\cite{li2023fast,wyatt2022anoddpm} and pixel-level guidance~\cite{kim2022diffusemorph,li2023fast,gong2023diffusion}, unconditional denoising networks can perform image translation across different types of scarce images, including high-quality format images, healthy images, and unbiased images. Notable examples include FNDM~\cite{li2023fast}, which enables accurate detection of brain anomalies through a non-Markovian framework with hybrid-condition guidance, and DiffuseMorph~\cite{kim2022diffusemorph}, which performs MR image registration using continuous diffusion sampling conditioned on moving and fixed image pairs.
Moreover, there are promising methods for enriching training datasets with realistic medical images generated from a small number of high-quality samples~\cite{pinaya2022brain,chung2022mr,dorjsembe2022three}. For instance, Latent Diffusion Models trained on 31,740 samples have been used to synthesize a high-quality and semantically rich dataset with 100,000 instances, achieving an impressive FID score of 0.0076~\cite{pinaya2022brain}.

\subsection{Text Generation}
Text generation plays a crucial role in bridging the gap between humans and advanced artificial intelligence by producing natural and coherent language. Autoregressive language models generate text sequentially, ensuring high semantic coherence but slower generation speed~\cite{otter2020survey}. On the other hand, diffusion models enable parallel text generation, offering faster speed but relatively weaker semantic coherence~\cite{zou2023diffusion,li2023diffusion}. Two primary approaches, namely \textbf{Discrete Generation} and \textbf{Latent Generation}, are commonly used to address the challenge of generating discrete tokens.

\subsubsection{Discrete Generation}
Discrete generation approaches involves models taking discrete words as input and utilize advanced techniques, parameterization, and pre-trained models. Pioneering the connection between diffusion models and discrete generation, typical works including D3PM~\cite{austin2021structured} and Argmax~\cite{hoogeboom2021argmax} treat words as categorical vectors. They establish forward and backward processes using a discrete transition matrix, considering the data to be generated as a stationary distribution. DiffusionBERT~\cite{he2022diffusionbert} combines diffusion models with pre-trained language models, showcasing improved text generation performance. Moreover, it introduces a novel noise schedule and explores the incorporation of time steps into BERT for reverse diffusion processes.

\subsubsection{Latent Generation}
The second approach focuses on generating text in the latent space of tokens, capturing the continuous nature of the diffusion process. It incorporates enhanced loss functions~\cite{yuan2022seqdiffuseq,wu2023ar,gao2022difformer}, diverse generation types~\cite{gong2022diffuseq,strudel2022self}, and advanced model architectures~\cite{li2022diffusion,lin2023text}. For instance, LM-Diffusion~\cite{li2022diffusion} introduces transformer-based graphical models for controllable generation, demonstrating superior performance in various text generation tasks. GENIE~\cite{lin2023text} presents a large-scale language model based on the diffusion framework, incorporating a novel Continuous Paragraph Denoise (CPD) loss for improved denoising and paragraph-level coherence. It showcases the potential of diffusion-based decoders for text generation and provides a strong foundation for future research.
In addition to advanced conditional sampling, token-level capturing, and post-refinement, diffusion models in NLP are expected to enhance the modeling of embedding space~\cite{strudel2022self,li2023diffusion}, establish connections with large pre-trained language models, and support cross-modality generations~\cite{tang2023learning,ye2023diffusion,zou2023diffusion}.

\subsection{Time Series Generation}
Accurate time series modeling is crucial for trend prediction, decision making, and real-time analysis. The diffusion model enhances this process with modules for time series data, enabling superior analysis and diverse generation \cite{lin2023diffusion}. Prior conditions can be categorized into inpainting tasks and prediction tasks based on different types of masking strategies. In inpainting tasks, observed states are used as prior conditions \cite{tashiro2021csdi,lopez2023diffusion,liu2023pristi,lim2023regular}, combined with context-based modules. CSDI proposed a self-supervised training framework based on bidirectional CNN modules, achieving substantial improvement in continuous generation of healthcare and environmental data \cite{tashiro2021csdi}. For prediction tasks, prior states are transformed into user-defined features and latent embeddings, serving as self-conditions \cite{rasul2021autoregressive,yan2021scoregrad,wen2023diffstg}. Combined with temporal-spatial modules, such as Graph UNet and RNN, DiffSTG and TimeGrad successfully achieve spatio-temporal probabilistic learning for time series \cite{wen2023diffstg,rasul2021autoregressive}. The success of time series generation hinges on the accurate modeling of time-dependent series and the incorporation of robust self-conditional guidance during sampling. These aspects point towards promising future advancements in the field \cite{fahim2023transfusion,lin2023diffusion,wen2023diffstg}.

\subsection{Audio Generation}
Synthesizing high-quality simulated speech has diverse applications in music composition, virtual reality, game development, and voice assistants, offering personalized and immersive audio experiences and improving human-computer interaction. Diffusion models, well-suited for handling the unique characteristics of audio data, utilize strong priors and effectively manage high-dimensional, time-dependent information.
Speech generation relies on hybrid conditions, combining text and control tags to achieve specific semantics or sound features. Techniques such as WaveGrad~\cite{chen2020wavegrad}, DiffSinger~\cite{liu2022diffsinger}, and DiffSVC~\cite{liu2021diffsvc} use Mel-Spectrogram as conditional guidance, while BinualGrad~\cite{leng2022binauralgrad} separates audio based on mono audio input. These methods form the foundation for general waveform generation, and additional features like loudness, melody, and phonetic posteriorgram enable controllable style generation~\cite{liu2021diffsvc,wu2021itotts,liu2022diffsinger,popov2021diffusion}.
Text-based and music-based generation, including text-to-speech and acoustic generation, rely on spectrogram features. Diffusion models incorporate text and rhythm as latent variables, leveraging spectrogram features and multi-view labels during sampling. Guided-TTS~\cite{kim2022guided} and Diff-TTS~\cite{jeong21_interspeech} employ components such as a speaker text encoder, duration predictor, and phoneme classifier for content generation and speech style guidance. Guide-TTS2~\cite{kim2022guided2} extends this approach to untranscribed speech generation using a classifier-free speaker encoder. Additional guidance factors include emotion, noise level, and music style~\cite{tae2021editts,koizumi2022specgrad,liu2022diffsinger,yang2022diffsound}.

\subsection{Molecule Design}
Molecules, as the fundamental building blocks of life, play a vital role in numerous biological processes. The design of functional molecules has long been a challenging and enduring problem \cite{min2017deep}. Generative models have revolutionized molecular design by offering a more efficient alternative to the traditional, laborious methods of enumeration and experimental validation. By characterizing specific modal distributions and functional domains, generative models can produce novel and effective drug molecule structures, expanding the possibilities in drug design \cite{bilodeau2022generative}. In the realm of drug discovery, diffusion models efficiently explore vast compound spaces, accelerating the search for potential drug candidates. This enhances the overall efficiency of the drug discovery process and reveals intricate compound relationships that contribute to a better understanding of drug mechanisms. The patterns observed in molecule design can be broadly categorized into unconditional generation and cross-modal generation.

\subsubsection{Unconditional Generation}
Unconditional molecule generation focuses on generating molecular structures using diffusion models, which offer speed and high-quality modeling capabilities. One approach is to generate the positions of molecules in three-dimensional space, capturing the conformation of molecules in space \cite{hoogeboom2022equivariant,xu2021geodiff,anand2022protein}. However, this approach may result in lower diversity and larger errors due to the non-uniform and irregular distribution of molecular three-dimensional structures. Alternatively, generating models that capture multiple features and the distribution of structural features in high-dimensional space can lead to more diverse distributions and interpretability \cite{lee2022proteinsgm,yim2023se,wu2022protein,jing2022torsional}. \cite{corso2023particle} further introduces a repulsion force between samples to promote the diveristy.

\subsubsection{Cross-modal Generation}
In molecular design, cross-modal generation focuses on incorporating functionality as a condition. Diffusion-based methods excel at incorporating conditions and leveraging denoising models based on different modalities to enhance modeling capabilities. Sequence-based cross-modal generation methods utilize protein sequences and multiple sequence alignments (MSA) sequences to train denoising models, incorporating specific protein structural information and functional labels to guide the generation \cite{lisanza2023joint,alamdari2023protein}. Structure-based cross-modal methods leverage prior knowledge from structure prediction models to assist in precisely guided generation, combining protein sequences and functional information \cite{watson2023novo}. Molecular docking and antibody design methods utilize the structural priors of target molecules to guide the docking process and identify favorable binding configurations \cite{corso2022diffdock,luo2022antigen}. These methods leverage the prior knowledge of target structures to enhance the generation and obtain promising conformations.

\subsection{Graph Generation}
The motivation for employing diffusion models to generate graphs stems from the aim to study and simulate diverse real-world networks and propagation processes. By doing so, it offers improved understanding and problem-solving capabilities for real-world issues. This approach empowers researchers to delve into the interactions and information propagation mechanisms within intricate systems, unveiling concealed patterns and correlations, and enabling the prediction of potential outcomes. The applications of this method encompass social network analysis, analysis of biological neural systems, as well as the generation and evaluation of graph datasets.
In \Secref{sec:new-space}, we have previously mentioned the conventional methods for graph generation, which involve generating an adjacency matrix or node features through discrete diffusion~\cite{huang2022graphgdp,vignac2022digress,niu2020permutation}. However, these unconditionally generated graphs have limited scalability and lack practical applicability. As a result, the predominant approach in graph generation revolves around generating graphs based on specific conditions and requirements.
Diffusion-based graph generation, guided by various specified conditions, facilitates the expansion of graph scale, refinement of graph features, and resolution of dataset-specific issues. PCFI~\cite{um2022confidence} leverages partial graph features and utilizes shortest path distances to predict pseudo confidence, serving as a guiding factor in the generation process. EDGE~\cite{chen2023efficient} and DiffFormer~\cite{wu2022difformer}, on the other hand, utilize node degree and energy constraints, respectively, as conditions to enable discrete and continuous generation of adjacency matrices and latent embeddings, thereby broadening the range of generation possibilities. Moreover, D4Explainer~\cite{chen2023d4explainer} incorporates the distribution of graph data as a condition and combines distribution loss and counterfactual loss to explore counterfactual instances.

\section{Conclusions \& Discussions}
\label{sec: conclusion}

\subsection{Conclusions}
The diffusion model becomes increasingly crucial to fields of deep learning. To utilize the power of the diffusion model, this paper provides a comprehensive and up-to-date review of several aspects of diffusion models using detailed insights on various attitudes, including fundamental theories, improved algorithms, and applications. We aspire for this survey to serve as a comprehensive guide for readers, elucidating the advancements in diffusion model enhancement and offering valuable insights into its practical applications.

\subsection{Comparison to Existing Surveys}

There is several existing surveys in the field of diffusion model, including general survey~\cite{yang2022diffusion}, survey in diverse fields including vision~\cite{croitoru2023diffusion}, language processing~\cite{zou2023diffusion,li2023diffusion}, audio~\cite{zhang2023audio}, time series~\cite{lin2023diffusion}, medical analysis~\cite{kazerouni2022diffusion}, and bioinformatics~\cite{guo2023diffusion,zhang2023survey}, and surveys in diverse data structures~\cite{fan2023generative,koo2023acomprehensive}. Compared to existing surveys, we conduct a comprehensive review with insights to broadly include algorithm enhancement and wide-range applications. Furthermore, we keep up-to-date updates of this field to track the latest improvements and maintain our GitHub Repository monthly for long-lasting analysis.

\subsection{Limitations and Future Directions}

\subsubsection{Challenges Under Data Limitations}
Except low inference speed, diffusion models often encounter difficulties in discerning patterns and regularities from low-quality data, leading to their inability to generalize to new scenarios or datasets. Additionally, handling large-scale datasets presents computational challenges such as extended training times, excessive memory usage, or failure to converge to the desired states, thus limiting the model's scale and complexity. Moreover, biased or uneven data sampling can restrict the model's capacity to generate outputs that are adaptable across diverse domains or demographics.

\subsubsection{Controllable Distribution-based Generation}
Improving the model's ability to understand and generate samples within specific distributions is essential for achieving better generalization with limited data. By focusing on identifying patterns and correlations in the data, the model can generate samples that closely match the training data and meet specific requirements. This requires effective data sampling, utilization techniques, and optimizing model parameters and structures. Ultimately, this enhanced understanding allows for more controlled and precise generation, leading to improved generalization performance.

\subsubsection{Advanced Multi-modal Generation Leveraging LLMs}
The future direction of diffusion models entails the advancement of multi-modal generation through the integration of Large Language Models (LLMs). This integration enables the model to generate outputs that encompass a combination of text, images, and other modalities. By incorporating LLMs, the model's understanding of the interplay between different modalities is enhanced, resulting in outputs that are more diverse and realistic. Moreover, LLMs significantly enhance the efficiency of prompt-based generation by effectively leveraging the connections between text and other modalities. Additionally, LLMs act as a catalyst for improving the diffusion model's generation capabilities, expanding the range of domains in which it can generate modalities.

\subsubsection{Integration with Machine Learning Fields}
Combining diffusion models with traditional machine learning theories offers new opportunities for enhancing performance in various tasks. Semi-supervised learning is particularly valuable in addressing the inherent challenges of diffusion models, such as generalization, and enabling effective conditional generation even with limited data. By utilizing unlabeled data, it strengthens the diffusion models' ability to generalize and achieve desirable performance when generating samples under specific conditions.

Furthermore, reinforcement learning plays a crucial role by employing fine-tuning algorithms to provide targeted guidance during the model's sampling process. This guidance ensures focused exploration and facilitates controlled generation. Additionally, incorporating additional feedback enriches reinforcement learning, leading to improved controllable conditional generation capabilities of the model.

\ifCLASSOPTIONcompsoc
  \section*{Acknowledgments}
\else
  \section*{Acknowledgment}
\fi

This work described in this paper was partially supported by a grant from the Research Grants Council of the Hong Kong Special Administrative Region, China 
(Project Number: T45-401/22-N)  and by a grant from the Hong Kong Innovation and Technology Fund (Project Number: ITS/241/21). This work is partially supported by the National Key Research and Development Program of China (No. 2022YFE0200700), the National Natural Science Foundation of China (No. 62006219 and 62376254), the Natural Science Foundation of Guangdong Province (No. 2022A1515011579). This work is also partially supported by the Science and Technology Innovation 2030 - Major Project (No. 2021ZD0150100) and the National Natural Science Foundation of China (No. U21A20427). 

\ifCLASSOPTIONcaptionsoff
  \newpage
\fi

\bibliographystyle{IEEEtran}
\bibliography{ref}

\clearpage
\appendices

\section{Sampling Algorithms}

In this section, we provide a brief guide on current mainstream sampling methods. We divide them into two parts: unconditional sampling and conditional sampling. For unconditional sampling, we present the original sampling algorithms for three landmarks. For conditional sampling, we divide them into the labeled condition and the unlabeled condition. 

\subsection{Unconditional Sampling}

\subsubsection{Ancestral Sampling}

\begin{algorithm} 
	\caption{Ancestral Sampling \cite{ho2020denoising}} 
	\begin{algorithmic}
            \STATE ${x}_{T} \sim \mathcal{N}({0}, {I})$
		\FOR {$t = T, ..., 1$}  
            \STATE ${z} \sim \mathcal{N}({0}, {I})$
            \STATE ${x}_{t-1}=\frac{1}{\sqrt{\alpha_{t}}}\left({x}_{t}-\frac{1-\alpha_{t}}{\sqrt{1-\bar{\alpha}_{t}}} \epsilon_{\theta}\left({x}_{t}, t\right)\right)+\sigma_{t} {z}$ 
            \ENDFOR
            \RETURN $x_0$
	\end{algorithmic} 
\end{algorithm}

\subsubsection{Annealed Langevin Dynamics Sampling}

\begin{algorithm} 
	\caption{Annealed Langevin Dynamics Sampling \cite{song2019generative}} 
	\begin{algorithmic}
            \STATE Initialize $x_0$
		\FOR {$i = 1, ..., L$}  
            \STATE $\alpha_i$ $\leftarrow$ $\epsilon \cdot \sigma_i^2/\sigma_L^2$
                \FOR {$t = 1, ..., L$} 
                \STATE ${z}_t \sim \mathcal{N}({0}, {I})$
                \STATE $\tilde{\mathrm{x}}_{t} = \tilde{\mathrm{x}}_{t-1}+\frac{\alpha_{i}}{2} \mathrm{~s}_{\boldsymbol{\theta}}\left(\tilde{\mathrm{x}}_{t-1}, \sigma_{i}\right)+\sqrt{\alpha_{i}} \mathrm{z}_{t}$ 
                \ENDFOR
                \STATE $\tilde{x}_0$ $\leftarrow$ $\tilde{x}_T$
            \ENDFOR
            \RETURN $\tilde{x}_T$
	\end{algorithmic} 
\end{algorithm}

\subsubsection{Predictor-Corrector Sampling}

\begin{algorithm}
    \caption{Predictor-Corrector Sampling \cite{song2020score}}
    \begin{algorithmic}
    \STATE ${x}_{N} \sim \mathcal{N}\left({0}, \sigma_{\max }^{2}{I}\right)$
    \FOR{$i=N-1$ to $0$}
    \STATE ${z} \sim \mathcal{N}({0}, {I})$ 
    \IF{Variance Exploding SDE}
    \STATE ${x}_{i}^{\prime} \leftarrow {x}_{i+1}+\left(\sigma_{i+1}^{2}-\sigma_{i}^{2}\right){s}_{\boldsymbol{\theta}} *\left({x}_{i+1}, \sigma_{i+1}\right)$
    \STATE ${x}_{i} \leftarrow {x}_{i}^{\prime}+\sqrt{\sigma_{i+1}^{2}-\sigma_{i}^{2}} {z}$
    \ELSIF{Variance Preserving SDE}
    \STATE ${x}_{i}^{\prime} \leftarrow\left(2-\sqrt{1-\beta_{i+1}}\right) {x}_{i+1}+\beta_{i+1} {s}_{\boldsymbol{\theta}} *\left({x}_{i+1}, i+1\right)$
    \STATE ${x}_{i} \leftarrow {x}_{i}^{\prime}+\sqrt{\beta_{i+1}} {z} \quad$
    \ENDIF
    \FOR{$j=1$ to $M$}
    \STATE ${z} \sim \mathcal{N}({0}, {I})$
    \STATE $\quad {x}_{i} \leftarrow {x}_{i}+\epsilon_{i} {s}_{\boldsymbol{\theta}} *\left({x}_{i}, \sigma_{i}\right)+\sqrt{2 \epsilon_{i}} {z}$ \\
    \ENDFOR
    \ENDFOR
    \RETURN ${x}_{0}$
    \end{algorithmic}
\end{algorithm}

\subsection{Conditional Sampling}

\subsubsection{Labeled Condition}

\begin{algorithm} 
	\caption{Classifier-guided Diffusion Sampling \cite{dhariwal2021diffusion}} 
	\begin{algorithmic}
            \STATE \textbf{Input:} class label $y$, gradient scale $s$
            \STATE ${x}_{T} \sim \mathcal{N}({0}, {I})$
            \FOR {$t = T, ..., 1$}
            \IF{DDPM Sampling}
                \STATE $\mu, \Sigma \leftarrow \mu_{\theta}(x_t), \Sigma_{\theta}(x_t)$ 
                \STATE ${x}_{t-q} \leftarrow $ sample from $\mathcal{N}\left(\mu+s \Sigma \nabla_{x_{t}} \log p_{\phi}\left(y \mid x_{t}\right), \Sigma\right)$ 
            \ENDIF
            \IF{DDIM Sampling}
                \STATE $\hat{\epsilon} \leftarrow \epsilon_{\theta}\left(x_{t}\right)-\sqrt{1-\bar{\alpha}_{t}} \nabla_{x_{t}} \log p_{\phi}\left(y \mid x_{t}\right)$
                \STATE $x_{t-1} \leftarrow \sqrt{\bar{\alpha}_{t-1}}\left(\frac{x_{t}-\sqrt{1-\bar{\alpha}_{t}} \hat{\epsilon}}{\sqrt{\bar{\alpha}_{t}}}\right)+\sqrt{1-\bar{\alpha}_{t-1}} \hat{\epsilon}$
            \ENDIF
            \ENDFOR
            \RETURN $x_0$
	\end{algorithmic} 
\end{algorithm}

\begin{algorithm} 
	\caption{Classifier-free Guidance Sampling \cite{ho2022classifier}} 
	\begin{algorithmic}
            \STATE \textbf{Input:} guidance  $w$, conditioning $c$,  SNR  $\lambda_1, ..., \lambda_T$
            \STATE ${z} \sim \mathcal{N}({0}, {I})$
		\FOR {$t = 1, ..., T$}  
            \STATE $\tilde{\boldsymbol{\epsilon}}_{t}=(1+w) \boldsymbol{\epsilon}_{\theta}\left(\mathbf{z}_{t}, \mathbf{c}\right)-w \boldsymbol{\epsilon}_{\theta}\left(\mathbf{z}_{t}\right)$
            \STATE $\tilde{\mathbf{x}}_{t}=\left(\mathbf{z}_{t}-\sigma_{\lambda_{t}} \tilde{\boldsymbol{\epsilon}}_{t}\right) / \alpha_{\lambda_{t}}$
            \STATE $\mathbf{z}_{t+1} \sim \mathcal{N}\left(\tilde{\boldsymbol{\mu}}_{\lambda_{t+1} \mid \lambda_{t}}\left(\mathbf{z}_{t}, \tilde{\mathbf{x}}_{t}\right),\left(\tilde{\sigma}_{\lambda_{t+1} \mid \lambda_{t}}\right)^{1-v}\left(\sigma_{\lambda_{t} \mid \lambda_{t+1}}^{2}\right)^{v}\right)$
            \ENDFOR
            \RETURN $z_{T+1}$
	\end{algorithmic} 
\end{algorithm}

\subsection{Unlabeled Condition}

\begin{algorithm} 
	\caption{Self-guided Conditional Sampling \cite{hu2022self}} 
	\begin{algorithmic}
            \STATE \textbf{Input: } guidance $w$, annotation map $f_{\psi}, g_{\phi}$, dataset $\mathcal{D}$, label $\mathbf{k}$,  segmentation label $\mathbf{k}_s$, image guidance $\hat{\mathbf{k}}$
            \STATE ${x}_{T} \sim \mathcal{N}({0}, {I})$
		\FOR {$t = T, ..., 1$}  
            \STATE $z \sim \mathcal{N}({0}, {I})$
            \IF{Self Guidance}
            \STATE $\Tilde{\epsilon} \leftarrow (1-w) \boldsymbol{\epsilon}_{\theta}\left(\mathbf{x}_{t}, t\right)+w \boldsymbol{\epsilon}_{\theta}\left(\mathbf{x}_{t}, t ; f_{\psi}\left(g_{\phi}(\mathbf{x} ; \mathcal{D}) ; \mathcal{D}\right)\right)$
            \ELSIF{Self-Labeled Guidance}
            \STATE $\Tilde{\epsilon} \leftarrow \boldsymbol{\epsilon}_{\theta}\left(\mathbf{x}_{t}, \operatorname{concat}[t, \mathbf{k}]\right)$
            \ELSIF{Self-Boxed Guidance}
            \STATE $\Tilde{\epsilon} \leftarrow \boldsymbol{\epsilon}_{\theta}\left(\operatorname{concat}[\mathbf{x}_{t}, \mathbf{k}_s], \operatorname{concat}[t, \mathbf{k}]\right)$
            \ELSIF{Self-Segmented Guidance}
            \STATE $\Tilde{\epsilon} \leftarrow \boldsymbol{\epsilon}_{\theta}\left(\operatorname{concat}[\mathbf{x}_{t}, \mathbf{k}_s], \operatorname{concat}[t, \hat{\mathbf{k}}]\right)$
            \ENDIF
            \STATE ${x}_{t-1}=\frac{1}{\sqrt{\alpha_{t}}}\left({x}_{t}-\frac{1-\alpha_{t}}{\sqrt{1-\bar{\alpha}_{t}}} \tilde{\epsilon}\right)+\sigma_{t} {z}$
            \ENDFOR
            \RETURN $x_0$
	\end{algorithmic} 
\end{algorithm}

\clearpage
\section{Evaluation Metric}

\subsection{Inception Score (IS)}
The inception score is built on valuing the diversity and resolution of generated images based on the ImageNet dataset \cite{borji2022pros,salimans2016improved}. It can be divided into two parts: diversity measurement and quality measurement. Diversity measurement denoted by $p_{IS}$ is calculated w.r.t. the class entropy of generated samples: the larger the entropy is, the more diverse the samples will be. Quality measurement denoted by $q_{IS}$ is computed through the similarity between a sample and the related class images using entropy. It is because the samples will enjoy high resolution if they are closer to the specific class of images in the ImageNet dataset. Thus, to lower $q_{IS}$ and higher $p_{IS}$, the KL divergence \cite{kullback1997information} is applied to inception score calculation:
\begin{equation}
\begin{aligned}
IS &= D_{K L}(p_{IS}\parallel q_{IS}) \\
&=\mathbb{E}_{x \sim p_{IS}}\left[\log \frac{p_{IS}}{q_{IS}}\right] \\
& =\mathbb{E}_{x \sim p_{IS}}[\log (p_{IS})-\log (p_{IS})]
\end{aligned}
\end{equation}

\vspace{-1mm}

\subsection{Frechet Inception Distance (FID)}
Although there are reasonable evaluation techniques in the Inception Score, the establishment is based on a specific dataset with 1000 classes and a trained network that consists of randomness such as initial weights, and code framework. Thus, the bias between ImageNet and real-world images may cause an inaccurate outcome. 

FID is proposed to solve the bias from the specific reference datasets. The score shows the distance between real-world data distribution and the generated samples using the mean and the covariance.
\begin{equation}
\mathrm{FID}=\left\|\mu_{r}-\mu_{g}\right\|^{2}+\operatorname{Tr}\left(\Sigma_{r}+\Sigma_{g}-2\left(\Sigma_{r} \Sigma_{g}\right)^{1 / 2}\right)
\end{equation}

where $\mu_g, \Sigma_g$ are the mean and covariance of generated samples, and $\mu_r, \Sigma_r$ are the mean and covariance of real-world data. 

\vspace{-1mm}

\subsection{Negative Log Likelihood (NLL)}

According to Razavi \textit{et al.}, \cite{razavi2019generating} negative log-likelihood is seen as a common evaluation metric that describes all modes of data distribution. Some diffusion models like improved DDPM \cite{nichol2021improved} regard the NLL as measurement of distribution matching.
\begin{equation}
\mathrm{NLL} = \mathbb{E}\left[-\log p_{\theta}\left({x}\right)\right]
\end{equation}

\newpage
\section{Benchmarks}
The benchmarks of landmark models along with improved techniques corresponding to \textbf{FID score}, \textbf{Inception Score}, and \textbf{NLL} are provided on diverse datasets which includes CIFAR-10 \cite{krizhevsky2009learning}, ImageNet\cite{krizhevsky2017imagenet}, and CelebA-64 \cite{liu2015faceattributes}. The selected performance are listed according to NFE in descending order to compare for easier access.

\subsection{Benchmarks on CelebA-64}
\begin{table}[htb]
\centering
\caption{Benchmarks on CelebA-64}
\setlength{\tabcolsep}{2.5mm}{
\begin{tabular}{l|p{1.2cm} p{1.2cm} p{1.2cm}}
\hline
Method & NFE & FID & NLL \\
\hline
NPR-DDIM \cite{bao2022estimating} & 1000 & 3.15 & -  \\
SN-DDIM \cite{bao2022estimating} & 1000 & 2.90 & -  \\
NCSN \cite{song2019generative} & 1000 & 10.23 & - \\
NCSN ++ \cite{song2020score} & 1000 & 1.92 & 1.97 \\
DDPM ++ \cite{song2020score} & 1000 & 1.90 & 2.10 \\ 
DiffuseVAE \cite{https://doi.org/10.48550/arxiv.2201.00308} & 1000 & 4.76 & - \\
Analytic DPM \cite{bao2022analytic} & 1000 & - & 2.66 \\
\hline 
ES-DDPM \cite{lyu2022accelerating} & 200 & 2.55 & - \\
PNDM \cite{liu2022pseudo} & 200 & 2.71 & - \\
\hline
ES-DDPM \cite{lyu2022accelerating} & 100 & 3.01 & - \\
PNDM \cite{liu2022pseudo} & 100 & 2.81 & - \\
Analytic DPM \cite{bao2022analytic} & 100 & - & 2.66 \\
NPR-DDIM \cite{bao2022estimating} & 100 & 4.27 & -  \\
SN-DDIM \cite{bao2022estimating} & 100 & 3.04 & -  \\
\hline 
ES-DDPM \cite{lyu2022accelerating} & 50 & 3.97 & - \\
PNDM \cite{liu2022pseudo} & 50 & 3.34 & - \\
NPR-DDIM \cite{bao2022estimating} & 50 & 6.04 & -  \\
SN-DDIM \cite{bao2022estimating} & 50 & 3.83 & -  \\
DPM-Solver Discrete \cite{lu2022dpm1} & 36 & 2.71 & - \\
\hline 
ES-DDPM \cite{lyu2022accelerating} & 20 & 4.90 & - \\
PNDM \cite{liu2022pseudo} & 20 & 5.51 & - \\
DPM-Solver Discrete \cite{lu2022dpm1} & 20 & 2.82 & - \\
\hline 
ES-DDPM \cite{lyu2022accelerating} & 10 & 6.44 & - \\
PNDM \cite{liu2022pseudo} & 10 & 7.71 & - \\
Analytic DPM \cite{bao2022analytic} & 10 & - & 2.97 \\
NPR-DDPM \cite{bao2022estimating} & 10 & 28.37 & -  \\
SN-DDPM \cite{bao2022estimating} & 10 & 20.60 & -  \\
NPR-DDIM \cite{bao2022estimating} & 10 & 14.98 & -  \\
SN-DDIM \cite{bao2022estimating} & 10 & 10.20 & -  \\
DPM-Solver Discrete \cite{lu2022dpm1} & 10 & 6.92 & - \\
\hline
ES-DDPM \cite{lyu2022accelerating} & 5 & 9.15 & - \\
PNDM \cite{liu2022pseudo} & 5 & 11.30 & -\\
\hline
\end{tabular}
}
\end{table}

\subsection{Benchmarks on ImageNet-64}

\begin{table}[htb]
    \caption{Benchmarks on ImageNet-64}
    \centering
    \setlength{\tabcolsep}{3mm}{
    \begin{tabular}{l|p{0.9cm} l l l}
    \hline
        Method & NFE & FID & IS & NLL \\ \hline
        MCG \cite{chung2022improving} & 1000 & 25.4 & - & - \\
        Analytic DPM \cite{bao2022analytic} & 1000 & - & - & 3.61 \\
        ES-DDPM \cite{lyu2022accelerating} & 900 & 2.07 & 55.29 & - \\ 
        Restart \cite{Xu2023RestartSF} & 623 & 1.36 & - & - \\\hline
        Efficient Sampling \cite{watson2021learning} & 256 & 3.87 & - & - \\ 
        Analytic DPM \cite{bao2022analytic} & 200 & - & - & 3.64 \\ 
        NPR-DDPM \cite{bao2022estimating} & 200 & 16.96 & - & - \\
        SN-DDPM \cite{bao2022estimating} & 200 & 16.61 & - & - \\
        ES-DDPM \cite{lyu2022accelerating} & 100 & 3.75 & 48.63 & - \\ \hline
        DPM-Solver Discrete \cite{lu2022dpm1} & 57 & 17.47 & - & - \\ 
        Restart \cite{Xu2023RestartSF} & 39 & 2.38 & - & - \\ 
        ES-DDPM \cite{lyu2022accelerating} & 25 & 3.75 & 48.63 & - \\ 
        GGDM \cite{watsonlearning} & 25 & 18.4 & 18.12 & - \\ 
        Analytic DPM \cite{bao2022analytic} & 25 & - & - & 3.83 \\
        NPR-DDPM \cite{bao2022estimating} & 25 & 28.27 & - & - \\
        SN-DDPM \cite{bao2022estimating} & 25 & 27.58 & - & - \\
        DPM-Solver Discrete \cite{lu2022dpm1} & 20 & 18.53 & - & - \\ \hline
        ES-DDPM \cite{lyu2022accelerating} & 10 & 3.93 & 48.81 & - \\ 
        GGDM \cite{watsonlearning} & 10 & 37.32 & 14.76 & - \\ 
        DPM-Solver Discrete \cite{lu2022dpm1} & 10 & 24.4 & - & - \\ 
        ES-DDPM \cite{lyu2022accelerating} & 5 & 4.25 & 48.04 & - \\ 
        GGDM \cite{watsonlearning} & 5 & 55.14 & 12.9 & - \\ \hline
    \end{tabular}
    }
\end{table}

\subsection{Benchmarks on CIFAR-10 Dataset}

\begin{table}[htb]
    \centering
    \caption{Benchmarks on CIFAR-10 (NFE $\geq$ 1000)}
    \begin{tabular}{l|p{1.0cm} p{1.0cm} l l}
    \hline
        Method & NFE & FID & IS & NLL \\ \hline
        Improved DDPM \cite{nichol2021improved} & 4000 & 2.90 & - & - \\ \hline
        VE SDE \cite{song2020score} & 2000 & 2.20 & 9.89 & - \\ 
        VP SDE \cite{song2020score} & 2000 & 2.41 & 9.68 & 3.13 \\
        sub-VP SDE \cite{song2020score} & 2000 & 2.41 & 9.57 & 2.92 \\ \hline
        DDPM \cite{ho2020denoising} & 1000 & 3.17 & 9.46 & 3.72 \\ 
        NCSN \cite{song2019generative} & 1000 & 25.32 & 8.87 & - \\
        SSM \cite{song2020sliced} & 1000 & 54.33 & - & - \\ 
        NCSNv2 \cite{song2020improved} & 1000 & 10.87 & 8.40 & - \\ 
        D3PM \cite{austin2021structured} & 1000 & 7.34 & 8.56 & 3.44 \\ 
        Efficient Sampling \cite{jolicoeur2021gotta} & 1000 & 2.94 & - & - \\
        NCSN++ \cite{song2020score} & 1000 & 2.33 & 10.11 & 3.04 \\ 
        DDPM++ \cite{song2020score} & 1000 & 2.47 & 9.78 & 2.91 \\ 
        TDPM \cite{zheng2022truncated} & 1000 & 3.07 & 9.24 & - \\ 
        VDM \cite{kingma2021variational} & 1000 & 4.00 & - & - \\ 
        DiffuseVAE \cite{https://doi.org/10.48550/arxiv.2201.00308} & 1000 & 8.72 & 8.63 & - \\ 
        Analytic DPM \cite{bao2022analytic} & 1000 & - & - & 3.59 \\ 
        NPR-DDPM \cite{bao2022estimating} & 1000 & 4.27 & - & - \\
        SN-DDPM \cite{bao2022estimating} & 1000 & 4.07 & - & - \\
        Gotta Go Fast VP \cite{zhang2022fast} & 1000 & 2.49 & - & - \\ 
        Gotta Go Fast VE \cite{zhang2022fast} & 1000 & 3.14 & - & - \\
        INDM \cite{kim2022maximum} & 1000 & 2.28 & - & 3.09 \\ \hline
    \end{tabular}
\end{table}

\begin{table}[ht]
    \centering
    \caption{Benchmarks on CIFAR-10 (NFE $<$ 1000)}
    \begin{tabular}{l|p{1.0cm} p{1.0cm} l l}
    \hline
        Method & NFE & FID & IS & NLL \\ \hline
        Diffusion Step \cite{https://doi.org/10.48550/arxiv.2206.05173} & 600 & 3.72 & - & - \\ 
        ES-DDPM \cite{lyu2022accelerating} & 600 & 3.17 & - & - \\ \hline
        Diffusion Step \cite{https://doi.org/10.48550/arxiv.2206.05173} & 400 & 14.38 & - & - \\ \hline
        Diffusion Step \cite{https://doi.org/10.48550/arxiv.2206.05173} & 200 & 5.44 & - & - \\ 
        NPR-DDPM \cite{bao2022estimating} & 200 & 4.10 & - & - \\
        SN-DDPM \cite{bao2022estimating} & 200 & 3.72 & - & - \\
        Gotta Go Fast VP \cite{zhang2022fast} & 180 & 2.44 & - & - \\ 
        Gotta Go Fast VE \cite{zhang2022fast} & 180 & 3.40 & - & - \\ 
        LSGM \cite{vahdat2021score} & 138 & 2.10 & - & - \\
        PFGM \cite{Xu2022PoissonFG} & 110 & 2.35 & - & - \\ \hline
        DDIM \cite{song2020denoising} & 100 & 4.16 & - & - \\ 
        FastDPM \cite{kong2021fast} & 100 & 2.86 & - & - \\ 
        TDPM \cite{zheng2022truncated} & 100 & 3.10 & 9.34 & - \\ 
        NPR-DDPM \cite{bao2022estimating} & 100 & 4.52 & - & - \\
        SN-DDPM \cite{bao2022estimating} & 100 & 3.83 & - & - \\
        DiffuseVAE \cite{https://doi.org/10.48550/arxiv.2201.00308} & 100 & 11.71 & 8.27 & - \\ 
        DiffFlow \cite{zhang2021diffusion} & 100 & 14.14 & - & 3.04 \\ 
        Analytic DPM \cite{bao2022analytic} & 100 & - & - & 3.59 \\ \hline
        Efficient Sampling \cite{jolicoeur2021gotta} & 64 & 3.08 & - & - \\ 
        DPM-Solver \cite{lu2022dpm1} & 51 & 2.59 & - & - \\ 
        DDIM \cite{song2020denoising} & 50 & 4.67 & - & - \\ 
        FastDPM \cite{kong2021fast} & 50 & 3.2 & - & - \\ 
        NPR-DDPM \cite{bao2022estimating} & 50 & 5.31 & - & - \\
        SN-DDPM \cite{bao2022estimating} & 50 & 4.17 & - & - \\
        Improved DDPM \cite{nichol2021improved} & 50 & 4.99 & - & - \\ 
        TDPM \cite{zheng2022truncated} & 50 & 3.3 & 9.22 & - \\ 
        DEIS \cite{jolicoeur2021gotta} & 50 & 2.57 & - & - \\ 
        gDDIM \cite{zhang2022gddim} & 50 & 2.28 & - & - \\ \hline
        DPM-Solver Discrete \cite{lu2022dpm1} & 44 & 3.48 & - & - \\ 
        STF \cite{Xu2023StableTF} & 35 & 1.90 & - & - \\
        EDM \cite{karras2022elucidating} & 35 & 1.79 & - & - \\
        PFGM++ \cite{Xu2023PFGMUT} & 35 & 1.74 & - & - \\ \hline
        Improved DDPM \cite{nichol2021improved} & 25 & 7.53 & - & - \\ 
        GGDM \cite{watsonlearning} & 25 & 4.25 & 9.19 & - \\ 
        NPR-DDPM \cite{bao2022estimating} & 25 & 7.99 & - & - \\
        SN-DDPM \cite{bao2022estimating} & 25 & 6.05 & - & - \\
        DDIM \cite{song2020denoising} & 20 & 6.84 & - & - \\ 
        FastDPM \cite{kong2021fast} & 20 & 5.05 & - & - \\ 
        DEIS \cite{jolicoeur2021gotta} & 20 & 2.86 & - & - \\ 
        DPM-Solver \cite{lu2022dpm1} & 20 & 2.87 & - & - \\ 
        DPM-Solver Discrete \cite{lu2022dpm1} & 20 & 3.72 & - & - \\ 
        Efficient Sampling \cite{jolicoeur2021gotta} & 16 & 3.41 & - & - \\ \hline
        NPR-DDPM \cite{bao2022estimating} & 10 & 19.94 & - & - \\
        SN-DDPM \cite{bao2022estimating} & 10 & 16.33 & - & - \\
        DDIM \cite{song2020denoising} & 10 & 13.36 & - & - \\ 
        FastDPM \cite{kong2021fast} & 10 & 9.90 & - & - \\ 
        GGDM \cite{watsonlearning} & 10 & 8.23 & 8.90 & - \\ 
        Analytic DPM \cite{bao2022analytic} & 10 & - & - & 4.11 \\ 
        DEIS \cite{jolicoeur2021gotta} & 10 & 4.17 & - & - \\
        DPM-Solver \cite{lu2022dpm1} & 10 & 6.96 & - & - \\ 
        DPM-Solver Discrete \cite{lu2022dpm1} & 10 & 10.16 & - & - \\ 
        Progressive Distillation \cite{salimans2022progressive} & 8 & 2.57 & - & - \\ 
        Denoising Diffusion GAN \cite{xiao2021tackling} & 8 & 4.36 & 9.43 & - \\ \hline
        GGDM \cite{watsonlearning} & 5 & 13.77 & 8.53 & - \\ 
        DEIS \cite{jolicoeur2021gotta} & 5 & 15.37 & - & - \\ 
        Progressive Distillation \cite{salimans2022progressive} & 4 & 3.00 & - & - \\ 
        TDPM \cite{zheng2022truncated} & 4 & 3.41 & 9.00 & - \\ 
        Denoising Diffusion GAN \cite{xiao2021tackling} & 4 & 3.75 & 9.63 & - \\ 
        Progressive Distillation \cite{salimans2022progressive} & 2 & 4.51 & - & - \\ 
        TDPM \cite{zheng2022truncated} & 2 & 4.47 & 8.97 & - \\ 
        Denoising Diffusion GAN \cite{xiao2021tackling} & 2 & 4.08 & 9.80 & - \\ 
        Denoising student \cite{luhman2021knowledge} & 1 & 9.36 & 8.36 & - \\ 
        Progressive Distillation \cite{salimans2022progressive} & 1 & 9.12 & - & - \\ 
        TDPM \cite{zheng2022truncated} & 1 & 8.91 & 8.65 & - \\ \hline
    \end{tabular}
\end{table}

\begin{table*}[!ht]
    \centering
    \caption{Details for Improved Diffusion Methods}
    \scalebox{0.98}{
    \begin{tabular}{l|l|l|l|l|l|l|l} 
    \hline
        Method & Year & Data & Model & Framework & Training & Sampling & Code \\ 
        \hline
        \multicolumn{8}{c}{\textbf{Landmark Works}} \\
        \hline
        DPM \cite{sohl2015deep} & 2015 & RGB Image & Discrete & Diffusion & $L_{simple}$ & Ancestral & \href{https://github.com/Sohl-Dickstein/Diffusion-Probabilistic-Models}{[code]} \\
        DDPM \cite{ho2020denoising} & 2020 & RGB Image & Discrete & Diffusion & $L_{simple}$ & Ancestral & \href{https://github.com/hojonathanho/diffusion}{[code]}\\ 
        NCSN \cite{song2019generative} & 2019 & RGB Image & Discrete & Score & $L_{DSM}$ &  Langevin dynamics & \href{https://github.com/ermongroup/ncsn}{[code]} \\ 
        NCSNv2 \cite{song2020improved} & 2020 & RGB Image & Discrete & Score & $L_{DSM}$ &  Langevin dynamics & \href{https://github.com/ermongroup/ncsnv2}{[code]} \\ 
        Score SDE \cite{song2020score} & 2020 & RGB Image & Continuous & SDE & $L_{DSM}$ & PC-Sampling & \href{https://github.com/yang-song/score_sde}{[code]} \\ 
        \hline 
        \multicolumn{8}{c}{\textbf{Improved Works}} \\
        \hline
        Progressive Distill \cite{salimans2022progressive} & 2022 & RGB Image & Discrete & Diffusion & $L_{simple}$ & DDIM Sampling & \href{https://github.com/google-research/google-research/tree/master/diffusion_distillation}{[code]}\\ 
        Denoising Student \cite{luhman2021knowledge} & 2021 & RGB Image & Discrete & Diffusion & $L_{Distill}$ & DDIM Sampling & \href{https://github.com/tcl9876/Denoising_Student}{[code]} \\ 
        TDPM \cite{zheng2022truncated} & 2022 & RGB Image & Discrete & Diffusion & $L_{DDPM\&GAN}$ & Ancestral & - \\ 
        ES-DDPM \cite{lyu2022accelerating} & 2022 & RGB Image & Discrete & Diffusion & $L_{DDPM\&VAE}$ & Conditional Sampling & \href{https://github.com/zhaoyanglyu/early_stopped_ddpm}{[code]} \\ 
        CCDF \cite{chung2022come} & 2021 & RGB Image & Discrete & SDE & $L_{simple}$ &  Langevin dynamics & \href{https://github.com/arpitbansal297/cold-diffusion-models}{[code]} \\
        Franzese's Model \cite{https://doi.org/10.48550/arxiv.2206.05173} & 2022 & RGB Image & Continuous & SDE & $L_{DSM}$ & DDIM Sampling & - \\ 
        FastDPM \cite{kong2021fast} & 2021 & RGB Image & Discrete & Diffusion & $L_{simple}$ & DDIM Sampling & \href{https://github.com/FengNiMa/FastDPM_pytorch}{[code]} \\ 
        Improved DDPM \cite{nichol2021improved} & 2021 & RGB Image & Discrete & Diffusion & $L_{hybrid}$ & Ancestral & \href{https://github.com/openai/improved-diffusion}{[code]} \\ 
        VDM \cite{kingma2021variational} & 2022 & RGB Image & Both & Diffusion & $L_{simple}$ & Ancestral & \href{https://github.com/google-research/vdm}{[code]} \\ 
        San-Roman's Model \cite{san2021noise} & 2021 & RGB Image & Discrete & Diffusion & $L_{DDPM\&Noise}$ & Ancestral & - \\ 
        Analytic-DPM \cite{bao2022analytic} & 2022 & RGB Image & Discrete & Score & $L_{Trajectory}$ & Ancestral & \href{https://github.com/baofff/Analytic-DPM}{[code]} \\ 
        NPR-DDPM \cite{bao2022estimating} & 2022 & RGB Image & Discrete & Diffusion & $L_{DDPM\&Noise}$ & Ancestral & \href{https://github.com/baofff/Extended-Analytic-DPM}{[code]} \\ 
        SN-DDPM \cite{bao2022estimating} & 2022 & RGB Image & Discrete & Score & $L_{square}$ & Ancestral & \href{https://github.com/baofff/Extended-Analytic-DPM}{[code]} \\ 
        DDIM \cite{song2020denoising} & 2021 & RGB Image & Discrete & Diffusion & $L_{simple}$ & DDIM Sampling & \href{https://github.com/ermongroup/ddim}{[code]} \\ 
        gDDIM \cite{zhang2022gddim} & 2022 & RGB Image & Continuous & SDE\&ODE & $L_{DSM}$ & PC-Sampling & \href{https://github.com/qsh-zh/gDDIM}{[code]} \\ 
        INDM \cite{kim2022maximum} & 2022 & RGB Image & Continuous & SDE & $L_{DDPM\&Flow}$ & PC-Sampling & - \\ 
        Gotta Go Fast \cite{zhang2022fast} & 2021 & RGB Image & Continuous & SDE & $L_{DSM}$ & Improved Euler & \href{https://github.com/AlexiaJM/score_sde_fast_sampling}{[code]} \\
        DPM-Solver \cite{lu2022dpm1} & 2022 & RGB Image & Continuous & ODE & $L_{DSM}$ & Higher ODE solvers & \href{https://github.com/luchengthu/dpm-solver}{[code]} \\
       Restart \cite{Xu2023RestartSF} & 2023 & RGB Image & Continuous & SDE & $L_{DSM}$ & $2^{nd}$ Order Heun & \href{https://github.com/Newbeeer/diffusion_restart_sampling}{[code]} \\
        EDM \cite{karras2022elucidating} & 2022 & RGB Image & Continuous & ODE & $L_{DSM}$ & $2^{nd}$ Order Heun & \href{https://github.com/crowsonkb/k-diffusion}{[code]} \\ 
        PFGM \cite{Xu2022PoissonFG} & 2022 & RGB Image & Continuous & ODE & $L_{DSM}$ & ODE-Solver & \href{https://github.com/Newbeeer/Poisson_flow}{[code]} \\ 
        PFGM++ \cite{Xu2023PFGMUT} & 2023 & RGB Image & Continuous & ODE & $L_{DSM}$ & $2^{nd}$ Order Heun & \href{https://github.com/Newbeeer/pfgmpp}{[code]} \\ 
        PNDM \cite{liu2022pseudo} & 2022 & Manifold & Discrete & ODE & $L_{simple}$ & Multi-step \& Runge-Kutta & \href{https://github.com/luping-liu/PNDM}{[code]}  \\ 
        DDSS \cite{watsonlearning}  & 2021 & RGB Image & Discrete & Diffusion & $L_{simple}$ & Dynamic Programming & - \\ 
        GGDM \cite{watson2021learning} & 2022 & RGB Image & Discrete & Diffusion & $L_{KID}$ & Dynamic Programming & - \\ 
        Diffusion GAN \cite{xiao2021tackling} & 2022 & RGB Image & Discrete & Diffusion & $L_{DDPM\&GAN}$ & Ancestral & \href{https://github.com/Zhendong-Wang/Diffusion-GAN}{[code]} \\ 
        DiffuseVAE \cite{https://doi.org/10.48550/arxiv.2201.00308} & 2022 & RGB Image & Discrete & Diffusion & $L_{DDPM\&VAE}$ & Ancestral & \href{https://github.com/kpandey008/DiffuseVAE}{[code]} \\
        DiffFlow \cite{zhang2021diffusion} & 2021 & RGB Image & Discrete & SDE & $L_{DSM}$ &  Langevin \& Flow Sampling & \href{https://github.com/qsh-zh/DiffFlow}{[code]} \\ 
        LSGM \cite{vahdat2021score} & 2021 & RGB Image & Continuous & ODE & $L_{DDPM\&VAE}$ & ODE-Slover & \href{https://github.com/NVlabs/LSGM}{[code]} \\
        Score-flow \cite{song2021maximum} & 2021 & Dequantization & Continuous & SDE & $L_{DSM}$ & PC-Sampling & \href{https://github.com/yang-song/score_flow}{[code]} \\ 
        PDM \cite{kim2021maximum} & 2022 & RGB Image & Continuous & SDE & $L_{Gap}$ & PC-Sampling & - \\ 
        ScoreEBM \cite{gao2020learning} & 2021 & RGB Image & Discrete & Score & $L_{Recovery}$ &  Langevin dynamics & \href{https://github.com/ruiqigao/recovery_likelihood}{[code]}  \\ 
        Song's Model \cite{song2021train} & 2021 & RGB Image & Discrete & Score & $L_{DSM}$ &  Langevin dynamics & - \\ 
        Huang's Model \cite{huang2021variational} & 2021 & RGB Image & Continuous & SDE & $L_{DSM}$ & SDE-Solver & \href{https://github.com/CW-Huang/sdeflow-light}{[code]}  \\ 
        De Bortoli's Model \cite{de2021simulating} & 2021 & RGB Image & Continuous & SDE & $L_{DSM}$ & Importance Sampling & \href{https://github.com/jeremyhengjm/DiffusionBridge}{[code]} \\ 
        PVD \cite{Zhou_2021_ICCV} & 2021 & Point Cloud & Discrete & Diffusion & $L_{simple}$ & Ancestral & \href{https://github.com/alexzhou907/PVD}{[code]} \\ 
        Luo's Model \cite{luo2021diffusion} & 2021 & Point Cloud & Discrete & Diffusion & $L_{simple}$ & Ancestral & \href{https://github.com/luost26/diffusion-point-cloud}{[code]} \\ 
        Lyu's Model \cite{lyu2021conditional} & 2022 & Point Cloud & Discrete & Diffusion & $L_{simple}$ & Farthest Point Sampling & \href{https://github.com/ZhaoyangLyu/Point_Diffusion_Refinement}{[code]} \\ 
        D3PM \cite{austin2021structured} & 2021 & Categorical Data & Discrete & Diffusion & $L_{hybrid}$ & Ancestral & \href{https://paperswithcode.com/paper/structured-denoising-diffusion-models-in}{[code]} \\ 
        Argmax \cite{hoogeboom2021argmax} & 2021 & Categorical Data & Discrete & Diffusion & $L_{DDPM\&Flow}$ & Gumbel sampling & \href{https://github.com/didriknielsen/argmax_flows}{[code]} \\ 
        ARDM \cite{hoogeboom2021autoregressive} & 2022 & Categorical Data & Discrete & Diffusion & $L_{simple}$ & Ancestral & \href{https://github.com/google-research/google-research/tree/master/autoregressive_diffusion}{[code]} \\ 
        Campbell's Model \cite{campbell2022continuous} & 2022 & Categorical Data & Continuous & Diffusion & $L_{simple}^{CT}$ & PC-Sampling & \href{https://github.com/andrew-cr/tauLDR}{[code]} \\ 
        VQ-diffusion \cite{gu2022vector} & 2022 & Vector-Quantized & Discrete & Diffusion & $L_{simple}$ & Ancestral & \href{https://github.com/microsoft/VQ-Diffusion}{[code]} \\ 
        Improved VQ-Diff \cite{tang2022improved} & 2022 & Vector-Quantized & Discrete & Diffusion & $L_{simple}$ & Purity Prior Sampling & \href{https://github.com/cientgu/VQ-Diffusion}{[code]} \\ 
        Cohen's Model \cite{cohen2022diffusion} & 2022 & Vector-Quantized & Discrete & Diffusion & $L_{simple}$ & Ancestral \& VAE Sampling & \href{https://github.com/maxjcohen/diffusion-bridges}{[code]} \\ 
        Xie's Model \cite{xie2022vector} & 2022 & Vector-Quantized & Discrete & Diffusion & $L_{DDPM\&Class}$ & Ancestral $\&$ VAE Sampling & - \\ 
        RGSM \cite{de2022riemannian} & 2022 & Manifold & Continuous & SDE & $L_{DSM}$ & Geodesic Random Walk & - \\ 
        RDM \cite{huang2022riemannian} & 2022 & Manifold & Continuous & SDE & $L_{simple}^{CT}$ & Importance Sampling & - \\ 
        EDP-GNN \cite{niu2020permutation} & 2020 & Graph & Discrete & Score & $L_{DSM}$ &  Langevin dynamics & \href{https://github.com/ermongroup/GraphScoreMatching}{[code]}  \\ 
        \hline
    \end{tabular}
    }
\end{table*}

\begin{table*}[!ht]
    \centering
    \caption{Details for Diffusion Applications}
    \scalebox{1.00}{
    \begin{tabular}{l|l|l|l|l|l}
    \hline
        Method & Year & Data & Framework & Downstream Task & Code \\ \hline
        \multicolumn{6}{c}{\textbf{Computer Vision}} \\ \hline
        CMDE \cite{batzolis2021conditional} & 2021 & RGB-Image & SDE & Inpainting,  Super-Resolution,  Edge to image translation & \href{conditional_score_diffusion}{[code]}  \\
        DDRM \cite{kawar2022denoising} & 2022 & RGB-Image & Diffusion & Super-Resolution, Deblurring, Inpainting, Colorization & \href{https://ddrm-ml.github.io/}{[code]}  \\ 
        Palette \cite{saharia2022palette} & 2022 & RGB-Image & Diffusion & Colorization, Inpainting, Uncropping, JPEG Restoration & \href{https://github.com/Janspiry/Palette-Image-to-Image-Diffusion-Models}{[code]} \\ 
        DiffC \cite{theis2022lossy} & 2022 & RGB-Image & SDE & Compression & - \\ 
        SRDiff \cite{li2022srdiff} & 2021 & RGB-Image & Diffusion & Super-Resolution - \\ 
        RePaint \cite{lugmayr2022repaint} & 2022 & RGB-Image & Diffusion & Inpainting,  Super-resolution,  Edge to Image Translation & \href{https://github.com/andreas128/RePaint}{[code]} \\ 
        FSDM \cite{giannone2022few} & 2022 & RGB-Image & Diffusion & Few-shot Generation & - \\ 
        CARD \cite{han2022card} & 2022 & RGB-Image & Diffusion & Conditional Generation & \href{https://github.com/xzwhan/card}{[code]} \\ 
        GLIDE \cite{nichol2021glide} & 2022 & RGB-Image & Diffusion & Conditional Generation & \href{https://github.com/openai/glide-text2im}{[code]} \\ 
        LSGM \cite{vahdat2021score} & 2022 & RGB-Image & SDE & UnConditional \& Conditional Generation & \href{https://github.com/NVlabs/LSGM}{[code]} \\ 
        SegDiff \cite{amit2021segdiff} & 2022 & RGB-Image & Diffusion & Segmentation & - \\ 
        VQ-Diffusion \cite{gu2022vector} & 2022 & VQ Data & Diffusion & Text-to-Image Synthesis & \href{https://github.com/microsoft/VQ-Diffusion}{[code]} \\ 
        DreamFusion \cite{poole2022dreamfusion} & 2023 & VQ Data & Diffusion & Text-to-Image Synthesis & \href{https://dreamfusion3d.github.io/}{[code]} \\ 
        Text-to-Sign VQ \cite{xie2022vector} & 2022 & VQ Data & Diffusion & Conditional Pose Generation & - \\ 
        Improved VQ-Diff \cite{tang2022improved} & 2022 & VQ Data & Diffusion & Text-to-Image Synthesis & - \\ 
        Luo's Model \cite{luo2021diffusion} & 2021 & Point Cloud & Diffusion & Point Cloud Generation & \href{https://github.com/luost26/diffusion-point-cloud}{[code]} \\ 
        PVD \cite{zhou20213d} & 2022 & Point Cloud & Diffusion & Point Cloud Generation, Point-Voxel representation & \href{https://github.com/alexzhou907/PVD}{[code]} \\ 
        PDR \cite{lyu2021conditional} & 2022 & Point Cloud & Diffusion & Point Cloud Completion & \href{https://github.com/ZhaoyangLyu/Point_Diffusion_Refinement}{[code]} \\ 
        Cheng's Model \cite{cheng2022autoregressive} & 2022 & Point Cloud & Diffusion & Point Cloud Generation & \href{https://github.com/AnjieCheng/CanonicalVAE}{[code]} \\ 
        Luo's Model\cite{luo2021score} & 2022 & Point Cloud & Score & Point Cloud Denoising & \href{https://github.com/AnjieCheng/CanonicalVAE}{[code]} \\ 
        VDM \cite{ho2022video} & 2022 & Video & Diffusion & Text-Conditioned Video Generation & \href{https://video-diffusion.github.io/}{[code]} \\ 
        RVD \cite{yang2022diffusion} & 2022 & Video & Diffusion & Video Forecasting, Video compression & \href{https://github.com/buggyyang/rvd}{[code]} \\ 
        FDM \cite{harvey2022flexible} & 2022 & Video & Diffusion & Video Forecasting, Long-range Video modeling & - \\ 
        MCVD \cite{voleti2022mcvd} & 2022 & Video & Diffusion & Video Prediction, Video Generation, Video Interpolation & \href{https://github.com/voletiv/mcvd-pytorch}{[code]} \\ 
        RaMViD \cite{hoppe2022diffusion} & 2022 & Video & SDE & Conditional Generation & - \\ 
        Score-MRI \cite{chung2022score} & 2022 & MRI & SDE & MRI Reconstruction & \href{https://github.com/HJ-harry/score-MRI}{[code]} \\ 
        Song's Model \cite{song2021solving} & 2022 & MRI, CT & SDE & MRI Reconstruction, CT Reconstruction & \href{https://github.com/yang-song/score_inverse_problems}{[code]} \\ 
        R2D2+ \cite{chung2022mr} & 2022 & MRI & SDE & MRI Denoising & - \\ \hline
        \multicolumn{6}{c}{\textbf{Sequence Modeling}} \\ \hline
        Diffusion-LM \cite{li2022diffusion} & 2022 & Text & Diffusion & Conditional Text Generation & \href{https://github.com/XiangLi1999/Diffusion-LM}{[code]} \\ 
        Bit Diffusion \cite{chen2022analog} & 2022 & Text & Diffusion & Image-Conditional Text Generation & \href{https://github.com/lucidrains/bit-diffusion}{[code]} \\ 
        D3PM \cite{austin2021structured} & 2021 & Text & Diffusion & Text Generation & - \\ 
        Argmax \cite{hoogeboom2021argmax} & 2021 & Text & Diffusion & Test Segmantation, Text Generation & \href{https://github.com/didriknielsen/argmax_flows}{[code]} \\ 
        CSDI \cite{tashiro2021csdi} & 2021 & Time Series & Diffusion & Series Imputation & \href{https://github.com/ermongroup/CSDI}{[code]} \\ 
        SSSD \cite{alcaraz2022diffusion} & 2022 & Time Series & Diffusion & Series Imputation & \href{https://github.com/AI4HealthUOL/SSSD}{[code]} \\ 
        CSDE \cite{park2021neural} & 2022 & Time Series & SDE & Series Imputation, Series Predicton & - \\ \hline
        \multicolumn{6}{c}{\textbf{Audio \& Speech}} \\ \hline
        WaveGrad \cite{chen2020wavegrad} & 2020 & Audio & Diffusion & Conditional Wave Generation & \href{https://github.com/lmnt-com/wavegrad}{[code]}  \\ 
        DiffWave \cite{kong2020diffwave} & 2021 & Audio & Diffusion & Conditional \& Unconditional Wave Generation & \href{https://github.com/lmnt-com/diffwave}{[code]}  \\ 
        GradTTS \cite{popov2021grad} & 2021 & Audio & SDE & Wave Generation & \href{https://github.com/huawei-noah/Speech-Backbones}{[code]} \\ 
        Diff-TTS \cite{jeong21_interspeech} & 2021 & Audio & Diffusion & non-AR mel-Spectrogram Generation, Speech Synthesis & -  \\ 
        DiffVC \cite{popov2021diffusion} & 2022 & Audio & SDE & Voice conversion & \href{https://diffvc-fast-ml-solver.github.io/}{[code]} \\ 
        DiffSVC \cite{liu2021diffsvc} & 2022 & Audio & Diffusion & Voice Conversion & \href{https://github.com/liusongxiang/diffsvc}{[code]}  \\
        DiffSinger \cite{liu2022diffsinger} & 2022 & Audio & Diffusion & Singing Voice Synthesis & \href{https://github.com/keonlee9420/DiffSinger}{[code]} \\ 
        Diffsound \cite{yang2022diffsound} & 2021 & Audio & Diffusion & Text-to-sound Generation tasks & \href{https://github.com/yangdongchao/Text-to-sound-Synthesis}{[code]} \\ 
        EdiTTS \cite{tae2021editts} & 2022 & Audio & SDE & fine-grained pitch, content editing & \href{https://github.com/neosapience/editts}{[code]} \\ 
        Guided-TTS \cite{kim2022guided} & 2022 & Audio & SDE & Conditional Speech Generation & - \\ 
        Guided-TTS2 \cite{kim2022guided2} & 2022 & Audio & SDE & Conditional Speech Generation & - \\ 
        Levkovitch's Model \cite{levkovitch2022zero} & 2022 & Audio & SDE & Spectrograms-Voice Generation & \href{https://diffvc-fast-ml-solver.github.io/}{[code]} \\ 
        SpecGrad \cite{koizumi2022specgrad} & 2022 & Audio & Diffusion & Spectrograms-Voice Generation & \href{https://wavegrad.github.io/specgrad/}{[code]} \\ 
        ItoTTS \cite{wu2021itotts} & 2022 & Audio & SDE & Spectrograms-Voice Generation & - \\ 
        ProDiff \cite{huang2022prodiff} & 2022 & Audio & Diffusion & Text-to-Speech Synthesis & \href{https://github.com/Rongjiehuang/ProDiff}{[code]} \\ 
        BinauralGrad \cite{leng2022binauralgrad} & 2022 & Audio & Diffusion & Binaural Audio Synthesis & - \\ \hline
        \multicolumn{6}{c}{\textbf{AI For Science}} \\ \hline
        ConfGF \cite{shi2021learning} & 2021 & Molecular & Score & Conformation Generation & \href{https://github.com/DeepGraphLearning/ConfGF}{[code]} \\ 
        DGSM \cite{luo2021predicting} & 2022 & Molecular & Score & Conformation Generation, Sidechain Generation & - \\ 
        GeoDiff \cite{xu2021geodiff} & 2022 & Molecular & Diffusion & Conformation Generation & \href{https://github.com/MinkaiXu/GeoDiff}{[code]} \\ 
        EDM \cite{hoogeboom2022equivariant} & 2022 & Molecular & SDE & Conformation Generation & \href{https://github.com/ehoogeboom/e3_diffusion_for_molecules}{[code]} \\ 
        Torsional Diff \cite{jing2022torsional} & 2022 & Molecular & Diffusion & Molecular Generation & \href{https://github.com/gcorso/torsional-diffusion}{[code]} \\ 
        DiffDock \cite{corso2022diffdock} & 2022 & Molecular\&protein & Diffusion & Conformation Generation, molecular docking & \href{https://github.com/gcorso/DiffDock}{[code]} \\ 
        CDVAE \cite{xie2021crystal} & 2022 & Protein & Score & Periodic Material Generation & \href{https://github.com/txie-93/cdvae}{[code]} \\ 
        Luo's Model \cite{luo2022antigen} & 2022 & Protein & Diffusion & CDR Generation & - \\ 
        Anand's Model \cite{anand2022protein} & 2022 & Protein & Diffusion & Protein Sequence and Structure Generation & - \\ 
        ProteinSGM \cite{lee2022proteinsgm} & 2022 & Protein & SDE & de novo protein design & - \\ 
        DiffFolding \cite{wu2022protein} & 2022 & Protein & Diffusion & Protein Inverse Folding & \href{https://github.com/microsoft/foldingdiff}{[code]} \\ \hline
    \end{tabular}}
\end{table*}

\section{Details for Improvement Algorithms}

\section{Table of Notation}

\begin{table}[!ht]
\vspace{-5mm}
\begin{center}
\caption{Notions in Diffusion Systems}
\begin{math}
\setlength{\tabcolsep}{5mm}{
\begin{tabular}{ l|l }
\hline
\textbf{Notations} & \textbf{Descriptions} \\ 
\hline
T & Discrete total time steps \\
\hline
t & Random time t  \\
\hline
${z}_t$ & Random noise with normal distribution \\
\hline
$\epsilon$ & Random noise with normal distribution \\
\hline
$\mathcal{N}$ & Normal distribution \\
\hline
$\beta$ & Generalized process noise scale \\
\hline
$\beta_t$ & Variance scale coefficients \\
\hline
$\beta(t)$ & Continuous-time $\beta_t$ \\
\hline
$\sigma$ & Generalized process noise scale \\
\hline
$\sigma_t$ & Noise scale of perturbation \\
\hline
$\sigma(t)$ & Continuous-time $\sigma_t$ \\
\hline
$\alpha_t$ & Mean coefficient defined as 1 - $\beta_t$ \\
\hline
$\alpha(t)$ &  Continuous-time $\alpha_t$ \\
\hline
$\bar{\alpha}_t$ & Cumulative product of $\alpha_t$ \\
\hline
$\gamma(t)$ & Signal-to-Noise ratio \\
\hline
$\eta_t$ & Step size of annealed Langevin dynamics \\
\hline
$x$ & Unperturbed data distribution \\
\hline
$\tilde{x}$ & Perturbed data distribution \\
\hline
$x_0$ & Starting distribution of data \\  
\hline
$x_t$ & Diffused data at time t \\
\hline
$x^{'}_{t}$ & Partly diffused data at time t \\
\hline
$x_T$ & Random noise after diffusion \\
\hline
$F(x, \sigma)$ & Forward/Diffusion process \\
\hline
$R(x, \sigma)$ & Reverse/Denoised process \\
\hline
$F_t(x_t, \sigma_t)$ & Forward/Diffusion step at time t \\
\hline
$R_t(x_t, \sigma_t)$ & Reverse/Denoised step at time t \\
\hline
$q(x_t | x_{t-1})$ & DDPM forward step at time t \\ 
\hline
$p(x_{t-1} | x_t)$ & DDPM reverse step at time t\\ 
\hline
${f}(x, t)$ & Drift coefficient of SDE \\
\hline
$g(t)$ & Simplified diffusion coefficient of SDE \\
\hline
$\mathcal{D}(x, t)$ & Degrader at time t in Cold Diffusion \\ 
\hline
$\mathcal{R}(x, t)$ & Reconstructor at time t in Cold Diffusion \\
\hline
${w}, {\bar{w}}$ & Standard Wiener process \\
\hline 
$\nabla_{{x}} \log p_{t}({x})$ & Score function w.r.t ${x}$ \\
\hline
$\mu_{\theta}(x_t, t)$ & Mean coefficient of reversed step \\
\hline
$\Sigma_{\theta}(x_t, t)$ & Variance coefficient of reversed step \\
\hline
$\epsilon_{\theta}(x_t, t)$ & Noise prediction model \\
\hline
$s_{\theta}({x})$ & Score network model \\
\hline
$L_0, L_{t-1}, L_T$ & Forward loss, reversed loss, decoder loss \\
\hline
$L_{vlb}$ & Evidence Lower Bound \\
\hline
$L_{vlb}^{CT}$ & Continuous evidence lower bound \\
\hline
$L_{simple}$ & Simplified denoised diffusion loss \\
\hline
$L_{simple}^{CT}$ & Continuous $L_{simple}$ \\
\hline
$L_{Gap}$ & Variational gap \\
\hline
$L_{KID}$ & Kernel inception distance \\
\hline
$L_{Recovery}$ & Recovery likelihood loss \\
\hline
$L_{hybrid}$ & Hybrid diffusion loss  \\
\hline
$L_{DDPM\&GAN}$ & DPM ELBO and GAN hybrid loss \\
\hline
$L_{DDPM\&VAE}$ & DPM ELBO and VAE hybrid loss\\
\hline
$L_{DDPM\&Flow}$ & DPM ELBO and normalizing flow hybrid loss\\
\hline
$L_{DSM}$ & Loss of denoised score matching \\
\hline
$L_{ISM}$ & Loss of implicit score matching \\
\hline
$L_{SSM}$ & Loss of sliced score matching \\
\hline
$L_{Distill}$ & Diffusion distillation loss \\
\hline
$L_{DDPM\&Noise}$ & DPM ELBO and reverse noise hybrid loss \\
\hline
$L_{Square}$ & Noise square loss \\
\hline
$L_{Trajectory}$ & Process optimization loss \\
\hline
$L_{DDPM\&Class}$ & DPM ELBO and classification hybrid loss \\
\hline
$\theta$ & learnable parameters \\
\hline
$\phi$ & learnable parameters \\
\hline
\end{tabular}
}
\end{math}
\end{center}
\vspace{-5mm}
\end{table}

\end{document}